\documentclass[lettersize,journal]{IEEEtran}
\usepackage{amsmath,amsfonts}
\usepackage{algorithm}
\usepackage{algpseudocode}
\usepackage{hyperref}
\usepackage{array}
\usepackage{courier}
\usepackage{xcolor}
\hypersetup{hidelinks}
\usepackage{subfigure}
\usepackage{textcomp,soul}
\usepackage{stfloats}
\usepackage{url}
\usepackage{verbatim}
\usepackage{graphicx}
\usepackage{booktabs}
\usepackage{multirow}
\usepackage{cite}
\usepackage{amsthm}
\usepackage{bm}
\usepackage{amssymb}
\theoremstyle{definition}

\newcommand{\MYhref}[3][blue]{\href{#2}{\color{#1}{#3}}}%
\makeatletter
\pretocmd\@bibitem{\color{black}\csname keycolor#1\endcsname}{}{\fail}
\newcommand\citecolor[1]{\@namedef{keycolor#1}{\color{blue}}}
\makeatother 
\citecolor{le2022algames}
\citecolor{zhu2023sequential}
\citecolor{peters2024contingency}
\citecolor{schwarting2021stochastic}
\citecolor{hua2022modeling}
\citecolor{wang2024a}
\citecolor{xu2021safe}
\citecolor{xu2018simulation}
\citecolor{hang2020integrated}
\citecolor{wang2019obstacle}
\citecolor{zhou2022autonomous}
\citecolor{wang2022driving}
\citecolor{lee2022design}
\citecolor{soto2022survey}

\title{CoDriveVLM: VLM-Enhanced Urban Cooperative Dispatching and Motion Planning for Future Autonomous Mobility on Demand Systems}
\author{Haichao Liu, Ruoyu Yao, Wenru Liu, Zhenmin Huang, Shaojie Shen, and Jun Ma
\thanks{Haichao Liu, Ruoyu Yao, and Wenru Liu are with the Robotics and Autonomous Systems Thrust, The Hong Kong University of Science and Technology (Guangzhou), China (e-mail: hliu369@connect.hkust-gz.edu.cn; ryao092@connect.hkust-gz.edu.cn; wliu354@connect.hkust-gz.edu.cn).}
\thanks{Zhenmin Haung, Shaojie Shen, and Jun Ma are with the Department of Electronic and Computer Engineering, The Hong Kong University of Science and Technology, Hong Kong SAR, China (e-mail: zhuangdf@connect.ust.hk;  eeshaojie@ust.hk; jun.ma@ust.hk).}
\thanks{This work has been submitted to the IEEE for possible publication.
Copyright may be transferred without notice, after which this version may
no longer be accessible.}}

\begin{document}
\maketitle
\begin{abstract}
The increasing demand for flexible and efficient urban transportation solutions has spotlighted the limitations of traditional Demand Responsive Transport (DRT) systems, particularly in accommodating diverse passenger needs and dynamic urban environments. Autonomous Mobility-on-Demand (AMoD) systems have emerged as a promising alternative, leveraging connected and autonomous vehicles (CAVs) to provide responsive and adaptable services. However, existing methods primarily focus on either vehicle scheduling or path planning, which often simplify complex urban layouts and neglect the necessity for simultaneous coordination and mutual avoidance among CAVs. This oversimplification poses significant challenges to the deployment of AMoD systems in real-world scenarios.
To address these gaps, we propose CoDriveVLM, a novel framework that integrates high-fidelity simultaneous dispatching and cooperative motion planning for future AMoD systems. Our method harnesses Vision-Language Models (VLMs) to enhance multi-modality information processing, and this enables comprehensive dispatching and collision risk evaluation. The VLM-enhanced CAV dispatching coordinator is introduced to effectively manage complex and unforeseen AMoD conditions, thus supporting efficient scheduling decision-making. Furthermore, we propose a scalable decentralized cooperative motion planning method via consensus alternating direction method of multipliers (ADMM) focusing on collision risk evaluation and decentralized trajectory optimization.
Simulation results demonstrate the feasibility and robustness of CoDriveVLM in various traffic conditions, showcasing its potential to significantly improve the fidelity and effectiveness of AMoD systems in future urban transportation networks. The code is available at \MYhref[black]{https://github.com/henryhcliu/CoDriveVLM.git}{https://github.com/henryhcliu/CoDriveVLM.git}. 

% Our framework represents a critical advancement towards realizing adaptable and reliable autonomous mobility solutions, addressing the limitations of existing methodologies and paving the way for intelligent urban mobility.
\end{abstract}
\begin{figure}[t]
    \centering
    \includegraphics[width=1\linewidth]{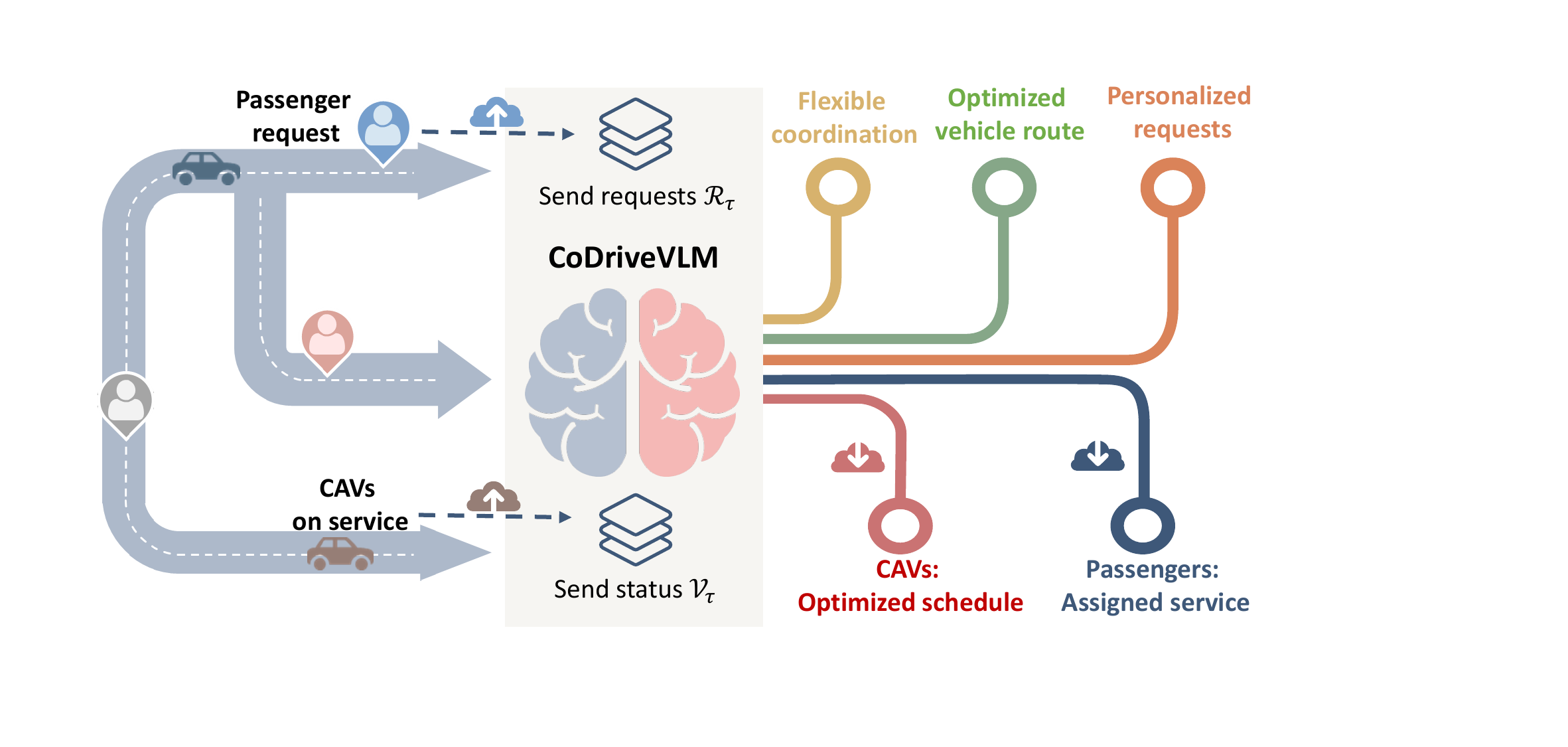}
    \caption{The dispatching of the CAVs for the passenger requests from the non-fixed taxi stops. All the requests are gradually assigned to the CAVs for personalized transportation services.}
    \label{fig:ProblemDemo}
\end{figure}
\section{Introduction}
Demand Responsive Transport (DRT) systems have been a cornerstone of public transportation strategies for several decades, offering a flexible alternative to traditional fixed-route services~\cite{shu2022novel,brake2004demand}. The core concept of DRT revolves around providing transportation services that can dynamically adapt to passenger demands, thereby optimizing resource use and enhancing user satisfaction. Initially, DRT was primarily associated with bus services to bridge the gap between fixed-route mass transit and individual taxi services~\cite{vansteenwegen2022survey}. Nonetheless, traditional DRT bus services have encountered several limitations, particularly in accommodating the diverse and dynamic transportation needs of modern urban environments,
% . These services often lack the flexibility needed to efficiently handle passengers with varying destinations and schedules, 
leading to inefficiencies and passenger dissatisfaction~\cite{wang2020joint}.
The advent of emerging technologies such as Vehicle-to-Everything (V2X)~\cite{chang2023bev,moubayed2020edge} and autonomous driving~\cite{hu2023planning,liu2023integrated} has revolutionized the landscape of DRT systems, which pave the way for more advanced and adaptable solutions. These innovations have significantly altered the operational framework of DRT, providing a technological foundation for the development of Autonomous Mobility-on-Demand (AMoD) systems~\cite{ding2023mechanism}. AMoD systems usually leverage the capabilities of connected and autonomous vehicles (CAVs) to offer highly responsive and efficient transportation services~\cite{paparella2024electric}. Essentially, AMoD systems promise to address the limitations of traditional DRT by enabling flexible coordination and optimization of vehicle routes and personalized passenger requests~\cite{horl2019dynamic}.

In AMoD systems, the cooperative driving of CAVs is a critical component, as it involves complex and dynamic task dispatching and path planning~\cite{guo2022rebalancing}. The scheduling of CAVs in response to passenger requests is particularly challenging due to the intricate configuration of urban traffic systems and diverse objectives of both passengers and CAVs. In detail, from the passengers' perspective, factors such as waiting time, total ride time, and the accessibility and reliability of CAV services are of paramount importance. Meanwhile, from the CAV management perspective, driving efficiency, safety, and effective cooperation among CAVs are essential to ensure seamless operations as well~\cite{fadleseed2024recent}. Therefore, designing a comprehensive scheduling strategy for the CAVs in AMoD systems is challenging because of the diverse objectives and spatial relationships that are difficult to model in the system~\cite{dandl2021regulating}. Recent advancements have explored the use of Reinforcement Learning (RL) for scheduling tasks within AMoD systems. A notable example involves the combination of multi-agent soft actor-critic with weighted bipartite matching to achieve scheduling on a smaller scale agents~\cite{enders2023hybrid}. Despite the innovative nature of these approaches, RL-based methods encounter significant challenges and the scenario-specific feature hinders their deployment in real and more widespread environments. Specifically, the inherent complexity of AMoD dispatching, characterized by numerous variables and scenarios, complicates the design of a clear and effective reward function~\cite{he2023robust}. In addition, the limited availability of real-world or simulated datasets for AMoD task scheduling hampers the practical applicability and training efficiency of data-driven models~\cite{iglesias2018data}.
In response to these challenges, various rule-based strategies have been proposed for task allocation, including Distance First, Idle First, Priority First, and so on~\cite{dandl2021regulating,guo2022rebalancing, chen2024llm}. These methods, structured around predefined rules, offer advantages in terms of simplicity and interoperability with low computational cost. However, due to the dynamic nature of urban transportation systems, they often fall short of providing the flexibility and adaptability required to achieve efficient scheduling. Algorithms designed for optimal assignment in transportation systems often model the problem using Mixed-Integer Linear Programming (MILP)~\cite{tsao2019model,fluri2019learning}. While this approach can yield precise and theoretically sound solutions for smaller-scale scenarios, it becomes increasingly impractical when deployed to larger and more complex systems. 
% This limitation poses a substantial challenge for real-time applications in AMoD systems, where swift decision-making is critical to maintaining operational efficiency and meeting passenger demands. Consequently, researchers are exploring alternative methodologies that can provide near-optimal solutions more efficiently, such as heuristics, metaheuristics, and machine learning-based approaches. These alternatives aim to strike a balance between solution quality and computational feasibility, thereby enabling the effective deployment of optimal assignment algorithms in large-scale AMoD environments.

On the other hand, the emergence of Large Language Models (LLMs) introduces new opportunities for task scheduling, leveraging their abilities in open-world understanding, reasoning, and few-shot in-context learning of common-sense knowledge~\cite{huang2023voxposer, yao2024tree}.
% zu2024language, }. 
Initiatives such as DiLu have capitalized on textual environmental descriptions for single-vehicle decision-making~\cite{wen2023dilu}. The ability of LLMs to make contextual inferences and decisions points to a promising future for AMoD scheduling systems. For instance, LiMeda is presented for multiple heterogeneous vehicles for different tasks for smart cities, which shows the potential of LLMs in solving vehicle dispatching problems for intelligent transportation systems~\cite{chen2024llm}. However, the application of multimodal LLMs, especially Vision-Language Models (VLMs), in AMoD scheduling and cooperative motion planning is still an emerging field with limited exploration~\cite{cui2024survey}. Given their strong capabilities in information representation, particularly in processing images and textual information, VLMs have the potential to significantly enhance the representation and understanding of urban transportation systems~\cite{liu2024lmmcodrive}, which highlight an important avenue for CAVs' dispatching research~\cite{choudhary2024talk2bev}.

Despite the potential of modern AMoD systems, current research efforts predominantly focus on vehicle scheduling, and the necessity of simultaneous motion planning and mutual avoidance among assigned vehicles after receiving the dispatching commands is overlooked. Many existing studies simplify the problem by treating maps or road layouts as grids~\cite{ma2013t,tong2018unified}, which fail to address the complexities encountered in real-world deployments. In other representative works, pre-calculated shortest paths and travel time between all nodes are stored in the road network~\cite{li2021optimal}. Thus travel time from a vehicle's starting point to its destination is directly retrieved from a predefined table without executing cooperative motion planning of the vehicles and potential route conflicts are neglected with ideal assumptions~\cite{alonso2017demand}. This simplification poses a significant challenge to the successful deployment of AMoD systems in complex urban environments. Furthermore, motion planning for AMoD systems is typically limited to simple scenarios, such as isolated road segments~\cite{karlsson2018multi,huang2024parallel}. For cooperative motion planning with predefined reference paths of the CAVs, learning-based and optimization-based methods are studied extensively following the scheduling phase within an AMoD system. Data-driven algorithms have gained attraction in addressing motion planning challenges for mobile robots and vehicles~\cite{teng2023motion,
% duan23relaxed,
liu2023towards}. In the realm of multi-agent pathfinding, PRIMAL utilizes imitation learning to mimic a centralized expert, enabling decentralized and efficient path planning while allowing for implicit agent coordination during execution~\cite{sartoretti2019primal}. However, despite its scalability for larger teams, PRIMAL's effectiveness diminishes in structured, densely populated environments that necessitate extensive coordination among agents~\cite{damani2021primal}. To overcome this drawback, RL-APCP$^3$ employs a bionic SARSA algorithm, inspired by slime mold behavior, to enhance performance in dense mapping scenarios~\cite{liu2021solving}. For CAVs, a learning-based iterative optimization approach has been developed to navigate collision-free cooperative motion planning at unsignalized intersections~\cite{klimke2022cooperative}. Additionally, a novel multi-agent behavioral planning strategy leveraging RL and graph neural networks has been proposed for urban intersections, significantly boosting vehicle throughput~\cite{wang2023coordination}. Nonetheless, learning-based methods often demand extensive real-world data and lack interpretability, which limits their applicability across diverse traffic scenarios~\cite{guo2023sustainability}. On the other hand, optimization-based methods for cooperative motion planning typically frame this task as an Optimal Control Problem (OCP). This framework provides a clear mathematical formulation and ensures optimality~\cite{li202optimization, huang2024parallel}. For CAVs characterized by nonlinear vehicle models, OCPs are typically tackled using established nonlinear programming solvers, such as the interior point optimizer (IPOPT). 
% Additionally, the iterative linear quadratic regulator~(iLQR) and differential dynamic programming~(DDP) address nonlinear optimization challenges by utilizing the Gauss-Newton approximation to maintain only the first-order term of agent dynamics~
Additionally, differential dynamic programming (DDP) addresses nonlinear OCPs by utilizing an approximation that considers first-order and second-order derivatives of the system dynamics\cite{lee2022gpu}, and the iterative linear quadratic regulator (iLQR) only considers the first-order terms via Gauss-Newton approximation to alleviate computational burden. However, standard DDP and iLQR methods are not designed to directly handle inequality constraints without additional modifications or constraint-handling techniques. Therefore, advanced versions like control-limited DDP and constrained iLQR have been developed~\cite{chen2019autonomous,Ma2022Alternating,ma2023local}.
To mitigate the computational demands associated with cooperative driving in expansive AMoD systems, the alternating direction method of multipliers (ADMM) presents an effective approach by breaking down the original optimization problem into a series of sub-problems. The distributed nature of ADMM makes it especially well-suited for cooperative motion planning among CAVs. For example, dual consensus ADMM has been implemented to establish a fully parallel optimization framework for the cooperative motion planning of CAVs~\cite{huang2023decentralized}. This strategy effectively distributes computational responsibilities across all participating entities, achieving real-time performance. Despite these advancements, existing methodologies struggle to support large-scale cooperative driving due to the fully connected nature of the agents. A recent work that exploits the sparsity feature of optimization problems for involved vehicles~\cite{liu2024improved}, indicates a promising direction for enhancing computational efficiency in AMoD systems. However, the vehicle grouping strategy is still oversimplified, and this highlights another critical research gap that remains to be addressed.

With the above discussion as a backdrop, designing and executing a cooperative driving scheme that effectively supports passenger requests in future AMoD systems presents several challenges. First, achieving seamless integration of scheduling and collaborative motion planning for high-fidelity simulation effects, which are essential for deploying algorithms in future intelligent transportation systems. Second, robust and swift vehicle collaborative scheduling to handle complex road structures and multi-task multi-objective demand scenarios. Third, collaborative motion planning of large-scale CAVs facilitating dynamic request assignments, which requires the development of real-time guiding path generation and trajectory updating to adjust tasks as needed dynamically and safely.

To address the aforementioned challenges, we present CoDriveVLM, a VLM-enabled urban cooperative driving framework. Our main contributions are as follows:
\begin{itemize}
    \item We formulate a high-fidelity simultaneous dispatching and cooperative motion planning platform that integrates multi-modality information generation, which enables the CoDriveVLM to operate dispatching and motion planning tasks cohesively.
    \item We introduce the VLM-enhanced CAV dispatching coordinator with few-shot learning support, which is capable of managing complex and unseen AMoD conditions, facilitating efficient scheduling decision-making for CAVs.
    \item In response to the dynamic dispatching requirements, we propose a VLM-ADMM-hybrid system for cooperative motion planning, focusing on collision risk evaluation and efficient decentralized trajectory optimization.
    \item The simulations in CARLA demonstrate the feasibility of our proposed high-fidelity AMoD simulation platform, and the results highlight the effectiveness and potential of VLM-enhanced decision-making modules for future transportation systems.
\end{itemize}

The rest of the paper is organized as follows: Section II provides the mathematical notations in the paper. In Section III, we formulate the problem and introduce the architecture of the AMoD system employed in the paper. In Section IV and Section V, we present a closed-loop cooperative dispatching and motion planning strategy supported by the CoDriveVLM framework. In Section VI, we demonstrate the performance of the proposed methodology through a series of simulations. Section VII presents the conclusion.

\section{Notations}
In this paper, we use $\mathcal{O}$ to express a multi-channel BEV image with different kinds of semantic objects, e.g., occupied and free vehicles, road layouts, lane centers, and passenger requests. The corresponding language-enhanced map of $\mathcal{O}$ is denoted as $L(\mathcal{O})$, which contains the textual semantic description and geometric cues for further implementation in the prompt engineering for VLMs. We define an undirected graph $\mathcal{G} = (\mathcal{N}, \mathcal{E})$, where $\mathcal{N}$ denotes the set of nodes and $\mathcal{E}$ represents the set of edges between these pairs of nodes. The subgraphs of the graph $ \mathcal{G} $ are denoted as $ \mathcal{H} $.
% and evolve over the cooperative driving planning horizons. Notably, vehicle $ n^i $ shares information only with neighboring vehicles within a specified communication range.
The number of CAVs in the $k$th subgraph $ \mathcal{H}_k = (\mathcal{N}_k, \mathcal{E}_{k,h}) $ is denoted as $ N_k $, where the index $k$ of the subgraph is omitted for clear representation in the rest of the paper. An edge $ (n^i, n^j) = d^{i,j} \in \mathcal{E}_h $ exists if the distance $ d^{i,j} $ is less than a threshold $ r^i_\text{tele} \in \mathbb{R}_+ $, where $\mathbb{R}_+$ is the set of non-negative real numbers. The cardinality of a node $ n^i $ in $ \mathcal{H} $ is defined as $ |{n^i}| = \text{deg}(n^i) $, representing the number of connected non-self nodes in the neighbor set:
\begin{equation}\label{eq:neighbor_node_set}
    \mathcal{N}^i = \{j \in \mathcal{N} \,|\, (i,j) \in \mathcal{E}_h\}.
\end{equation}
The set of non-negative integers is denoted by $\mathbb{Z}_+$. The set of logical or Boolean matrices is indicated by $\mathbb{B}$. The notation $[\, \cdot \,]^i$ corresponds to the variable associated with the $i$th CAV, and $[\, \cdot \,]_\tau$ indicates the variable at the time step $\tau$, e.g., $\mathcal{O}_\tau$ is the BEV image aggregated at the time step $\tau$. For clarity, we denote the row-concatenated vectors as $[\bm s^1, \bm s^2, \ldots, \bm s^n]$ and the column-concatenated matrix as $[\bm s^1; \bm s^2; \ldots; \bm s^n]$, where $\bm s^i$ is described as a state vector of the $i$th object. The squared weighted $L_2$-norm is expressed as $\bm x^\top \bm M \bm x$, which we simplify to $\| \bm x \|^2_{\bm M}$. An \textit{indicator function} with respect to the set $\mathbb{\mathcal{X}} \subseteq \mathbb{R}^n$ is represented as $\mathcal{I}_{\mathcal{X}}(\bm x)$, taking the value of 0 if $\bm x \in \mathcal{X}$ and $+\infty$ otherwise. The \textit{floor function}, denoted by $\lfloor x \rfloor$, is defined as the greatest non-negative integer in $\{n\in \mathbb{Z}_+\mid n \leq x < \left(n+1\right)\}$. 

\section{AMoD Framework Design and Formulation}

\subsection{Problem Analysis and Definition}

The demand for future transportation systems is rapidly evolving, driven by the increasing need for efficient, reliable, and personalized mobility solutions. AMoD systems are at the forefront of this evolution, promising to revolutionize urban transportation by leveraging CAVs. To realize the full potential of autonomous driving, these systems are recommended to facilitate unsignalized cooperative on-road driving (e.g., junction crossing), effectively navigating environments without traditional traffic lights to increase traffic flow. This necessitates advanced coordination and mutual avoidance strategies among CAVs to maintain high levels of safety and efficiency.

Moreover, personalized passenger service is a critical aspect of AMoD systems, as demonstrated by the launch of services such as Baidu Apollo Go, a leading robotaxi service in major cities such as Beijing and Wuhan~\cite{wang2024recent}. In such systems, requests are dynamically generated, and CAVs must respond promptly based on factors such as distance, passenger waiting time, and other contextual factors. The ability to accommodate a diverse range of passenger needs and preferences is paramount to the successful application of AMoD systems.

% \subsubsection{The Problems to be Solved in AMoD Systems}

Moreover, the primary challenges in AMoD systems involve the seamless integration of vehicle scheduling, path planning, and collision avoidance. Specifically, the system must efficiently assign CAVs to passenger requests, which are randomly generated on urban roads without fixed taxi stands or stops. This requires a robust dispatching mechanism capable of handling dynamic environments and fluctuating demand patterns.

Once assignments are made, the system must generate accurate navigation routes for CAVs to reach their destinations across different stages of the scheduling process. This involves determining optimal paths that consider traffic conditions, road networks, and passenger requirements.
Furthermore, handling collision avoidance is crucial as CAVs navigate urban roads to pick up and drop off passengers. Therefore, ensuring high commuting efficiency while preventing collisions demands a cooperative approach to motion planning, where CAVs communicate and coordinate with each other to optimize their movements.
Given the above analysis, we aim to design an integrated framework for AMoD systems that enables efficient, safe, and personalized urban transportation. The framework must address the following key challenges:

\begin{enumerate}
    \item \textbf{Dynamic Dispatching:} Efficiently assign CAVs to passenger requests, considering factors such as proximity, waiting time, and passenger preferences.
    
    \item \textbf{Optimal Routing:} Generate optimal travel routes for CAVs based on traffic conditions, road networks, and passenger needs to ensure efficient and timely service.
    
    \item \textbf{Collision Avoidance:} Implement coordination mechanisms to allow CAVs to safely navigate complex urban environments through cooperative motion planning and control.
    
    \item \textbf{Personalized Service:} Provide tailored services that meet the unique requirements and preferences of passengers to enhance user satisfaction.
\end{enumerate}

Our solution integrates elements from artificial intelligence, optimal control, and V2X communication to develop a comprehensive AMoD system capable of operating seamlessly in dynamic urban settings.

\subsection{Configuration Design of the AMoD System}\label{subsec:AmodStructure}

% \subsubsection{Architecture Levels}

Based on the above problem definition, the AMoD system architecture is designed around two primary sub-tasks: dispatching of the CAVs to the flexible passenger requests, and motion planning for the CAVs mitigating their route conflicts.

\subsubsection{Dispatching of the CAVs for Flexible Passenger Requests}

The dispatching process involves managing passenger requests and vehicle assignments. On the one hand, passenger requests are represented as a discrete mapping:
\begin{equation}\label{eq:def_setR}
    \mathcal{R} = \{ (j, \bm{d}_j, T_{\text{sp},j}, A_j, P_j, k_j, T_{\text{pk},j}, A_j, T_{\text{ar},j}) \}, j\in \mathbb{Z}_+,
\end{equation}
where $ j $ is the unique ID of a passenger request, $\bm{d}_j$ is the position vector $[x,y]^\top$ of the request, $T_{\text{sp},j}$ is the spawn time of the request $j$, $A_j,P_j,G_j\in \mathbb{B}$ are boolean indicating whether the request is assigned, picked and got to the destination, respectively. Besides, $k_j$ is the ID of the assigned CAV (or \texttt{None} if unassigned), $T_{\text{pk},j}$ is the pickup time, and $T_{\text{ar},j}$ is the arrival time of the same passenger $j$.

On the other hand, vehicle assignments are described as:
\begin{equation}\label{eq:def_setV}
\mathcal{V} = \{ (i, \bm{R}_i, h_i, F_i) \}, i\in \mathbb{Z}_+,
\end{equation}
where $ i $ is the specific ID of a CAV, $h_i=j$ is the assigned passenger request ID, $\bm{R}_i = [\bm r_0, \bm r_1, \ldots, \bm r_n]^\top$ is the reference path to the destination (location of the passenger request $j$ or its destination $\bm{d}_j$ in different stages), and $F_i\in \mathbb{B}$ is a boolean indicating the vehicle's availability.

\subsubsection{Motion Planning with Collision Mitigation}

For CAVs with $F_i = 1$, the task is to efficiently pick up and drop off assigned passengers $h_i$. This involves self-driving without collisions, necessitating cooperative motion planning and control strategies. Such strategies ensure that CAVs navigate complex urban environments while maintaining high efficiency and safety. 
Furthermore, following each receding horizon iteration of the motion planning process with a time period $T_p$, the terminal states of the CAVs are evaluated. 
Upon detecting that a task should transition to the next stage or conclude, the dispatching module executes the necessary procedures to ensure closed-loop and dynamic completion of autonomous mobility. 
The procedures involve updating the discrete mapping of the corresponding passenger request and CAV, re-planning the initial path for the CAV for subsequent time horizons.

\subsection{Coordination of the Frequency of the Sub-Systems}\label{subsec:hybridSystem}

The AMoD framework incorporates a hybrid system to balance the request needs of motion planning and vehicle dispatching. Trajectory planning with a receding horizon approach is crucial to adapt to dynamic changes. 
Conversely, dispatching can be less frequent when all CAVs are occupied and no new assignments are needed.
Based on the above consideration and inspired by~\cite{tian2024drivevlm}, we propose the hybrid system for the CoDriveVLM scheme that synergizes the strengths of VLMs with the optimization-based cooperative motion planning methods.

\subsubsection{Fast System}

The \textit{fast system} with a fixed frequency (e.g., a receding horizon of $T_p=2.0$\,s with a discrete time step of $\Delta T=0.1$\,s) focuses on cooperative motion planning and control, enabling CAVs to adjust their motion swiftly in response to real-time conditions in a closed-loop manner.
Concretely, within this structured framework, all CAVs collaboratively plan their trajectories based on the reference path $\bm{R}_i$ within the discrete mapping $\mathcal{V}$. This planning takes into account both mutual avoidance and overall driving efficiency, aiming to maintain the reference velocity $v_{\text{ref},i}$ as closely as possible. Therefore, this cooperative planning approach is supposed to ensure that CAVs operate harmoniously, minimizing potential conflicts and optimizing traffic flow.
The strategic coordination facilitated by this system is crucial for achieving seamless integration of CAVs within complex traffic environments, ultimately contributing to a more adaptive and responsive transportation network.
\subsubsection{Adaptive Frequency System}
The \textit{adaptive frequency system} manages the dispatching of free CAVs with $F_i = 1$, for newly added and pending passenger requests in the set $\{\bm{R}_j|k_j=\texttt{None}\}\in \mathcal{R}$. This system operates at a variable frequency, adapting to the availability of vehicles and the emergence of new requests over time. Specifically, the event-triggering condition is
\begin{equation}\label{eq:scheduleTriggering}
\left\{
\begin{aligned}
    &\{\bm{R}_j \mid k_j = \texttt{None}\} \neq \varnothing, & j \in \mathbb{Z}_+, \\
    &\{\bm{V}_i \mid F_i = 1\} \neq \varnothing, & i \in \mathbb{Z}_+,
\end{aligned}
\right.
\end{equation}
where $\bm{R}_j\in \mathcal{R}$ and $\bm{V}_i\in \mathcal{V}$, as the two sets are defined in (\ref{eq:def_setR}) and (\ref{eq:def_setV}). Therefore, once there are potential passenger requests for assignment, the dispatching module is executed to get a rational command to the available CAVs at that time step $\tau$.
By optimizing dispatching activity frequency in this manner, the system ensures efficient utilization of VLM inferring resources and enhances overall service responsiveness.

\section{Learnable VLM-enabled Cooperative Dispatching}

This section presents the VLM-enabled cooperative dispatching methodology, a core component of the CoDriveVLM framework. This methodology underpins the coordination of CAVs within AMoD systems, leveraging advanced multimodal data processing for optimal task assignment. The following content details the overall architecture, context generation, memory storage, and retrieval processes integral to this system.
\begin{figure*}[t]
    \centering
    \includegraphics[width=0.9\linewidth]{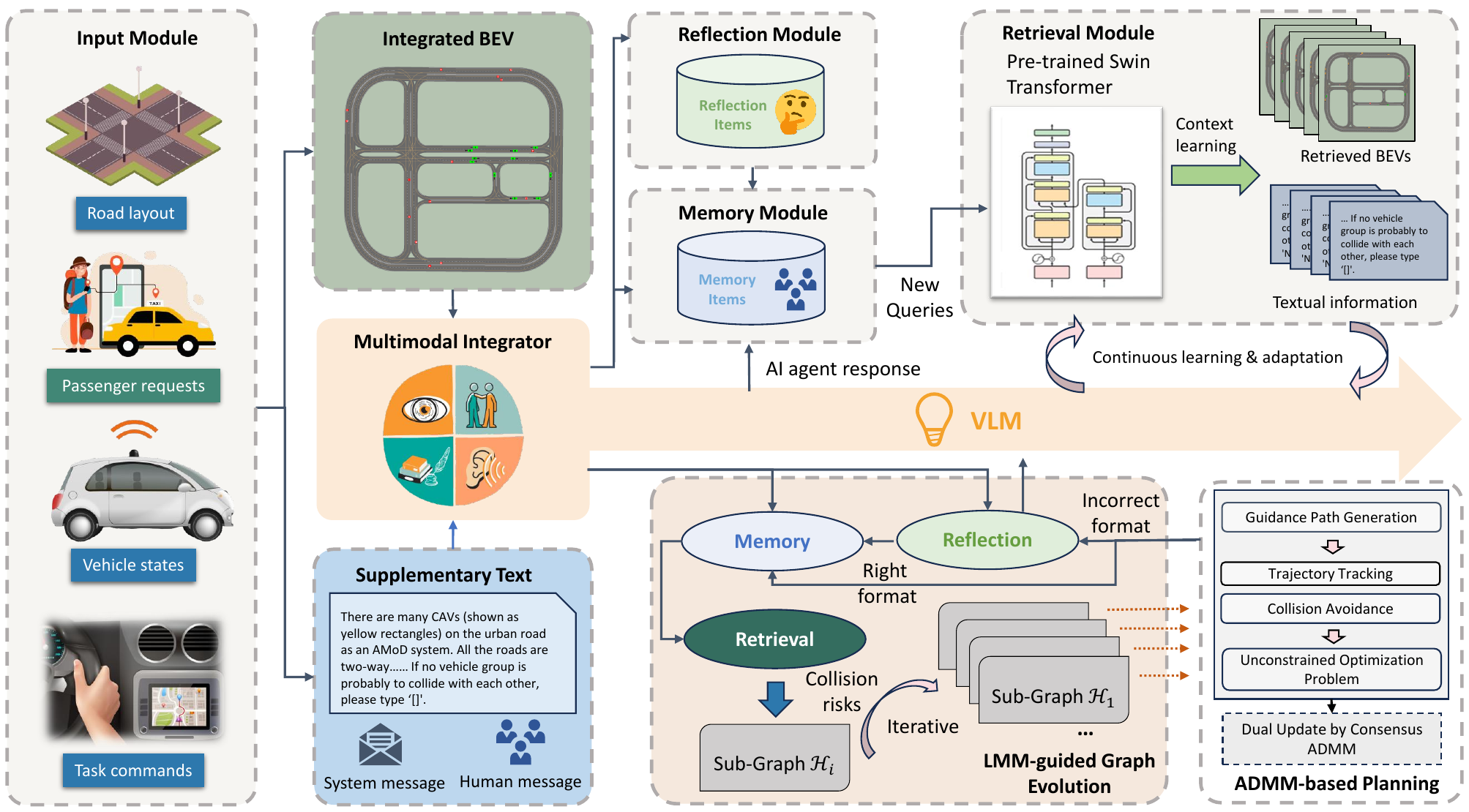}
    \caption{{Overall Architecture of the Proposed CoDriveVLM.} This framework encompasses multiple modules, from multimodal input processing to multifunctional output generation of the VLM. The input module consolidates information from traffic infrastructure, passengers, and CAVs for integration by the subsequent multimodal integrator. Within this framework, the VLM performs two primary functions: determining dispatch orders for CAVs in response to passenger requests, and assessing collision risks between CAV pairs to inform subgraph evolution for cooperative motion planning. Additionally, memory and reflection modules facilitate few-shot learning capabilities.}
    \label{fig:CoDriveVLMFramework}
\end{figure*}
\subsection{Overall Architecture}

The overarching architecture of the presented CoDriveVLM is based on a hybrid system, featuring different operating frequencies in its upper and lower components, as illustrated in Fig.~\ref{fig:CoDriveVLMFramework}.
Concretely, it contains the dispatching coordinator and the cooperative motion planner. Together, they facilitate closed-loop schedule and passenger transport operations, ensuring that each CAV is controlled with a microscopic precision that guarantees safety, efficiency, and keeps a good passenger experience.
The dispatching coordinator, enhanced by the context understanding and reasoning capabilities of VLMs, processes multimodal information and inferences with the CoT technique to assign tasks to CAVs dynamically. 

\subsection{Context Generator}

The context generator plays a crucial role in synthesizing perceptive elements from the urban environment, laying the groundwork for effective dispatching decisions. As marked in the first module of Fig.~\ref{fig:CoDriveVLMFramework}, it integrates information about road layouts, passenger requests, vehicle states, and others, producing a comprehensive situational awareness model.

\subsubsection{Perceptive Elements}

The perceptive elements denote the foundational data inputs for the context generator. They encompass the road layout, capturing the physical configuration of urban roads, including lanes, intersections, the center lines of the lanes, and so on. This data is critical for understanding the navigational constraints and opportunities within the environment. As expressed by $\mathcal{R}$ in (\ref{eq:def_setR}), passenger requests are also crucial, providing real-time data regarding passenger demand, including pickup and drop-off locations, which guide the task assignment and routing of CAVs. As depicted by $\mathcal{V}$ in (\ref{eq:def_setV}), vehicle states are represented as a discrete mapping of CAV positions, orientations, and operational statuses, facilitating the dynamic allocation of tasks. Lastly, the task commands, which describe the AMoD system's purpose, direct the VLMs' decision for the operational behavior of CAVs.

\subsubsection{Annotation Rules of Multimodal Contexts as VLM inputs}\label{subsubsec: annotationVLMDispatchingAgent}

The annotation rules are critical for transforming raw perceptive data into actionable insights. These rules dictate the representation of information within both BEV images and supplementary text.

\paragraph{BEV Image Annotation}
\begin{figure}[t]
    \centering
    \includegraphics[width=1\linewidth]{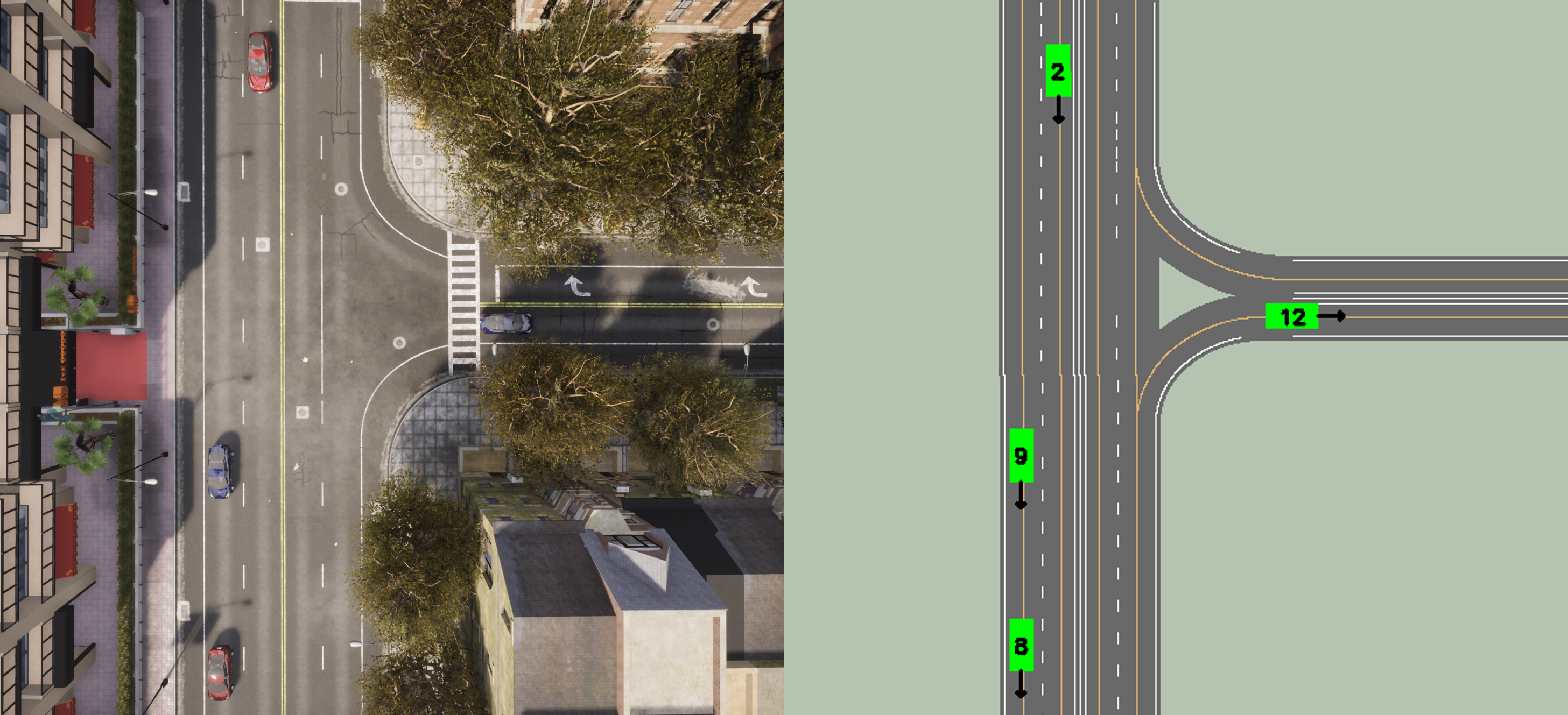}
    \caption{Annotation illustration of the BEV image. This image aggregates the information from the road layout, center lines of the lanes, and the states of the traffic participants. The arrows denote the driving direction of the CAVs, and the numbers above each object are their unique ID to facilitate language-based reasoning and inference.}
    \label{fig:bev_image_demo}
\end{figure}
One of the main reasons for using BEV images is to visually represent the road layout for the understanding of VLMs, as textual descriptions used in traditional LLMs can be ambiguous and insufficient. These images provide a precise spatial context for CAV operations and passenger request allocations. As described in Fig.~\ref{fig:bev_image_demo}, passenger requests are represented by red squares, while vehicles are depicted using color-coded rectangles, with green indicating available vehicles and yellow indicating those currently occupied. This visual differentiation aids in the quick assessment of fleet status. Arrows ``$\rightarrow$'' are placed in front of each vehicle to indicate its heading, facilitating understanding of vehicle movement patterns. The precise locations of passengers are also mapped in the BEV image with a fixed number of pixels per meter in the real AMoD system, enabling rational task allocation. To enhance the VLM's performance in reasoning tasks, unique ID numbers are assigned to each traffic object (CAVs and passenger requests), enabling clear reference and tracking. Until now, the multi-channel BEV image $\mathcal{O}$ is constructed and for further text information supplementary.
\begin{figure}[t]
    \centering
    \includegraphics[width=1\linewidth]{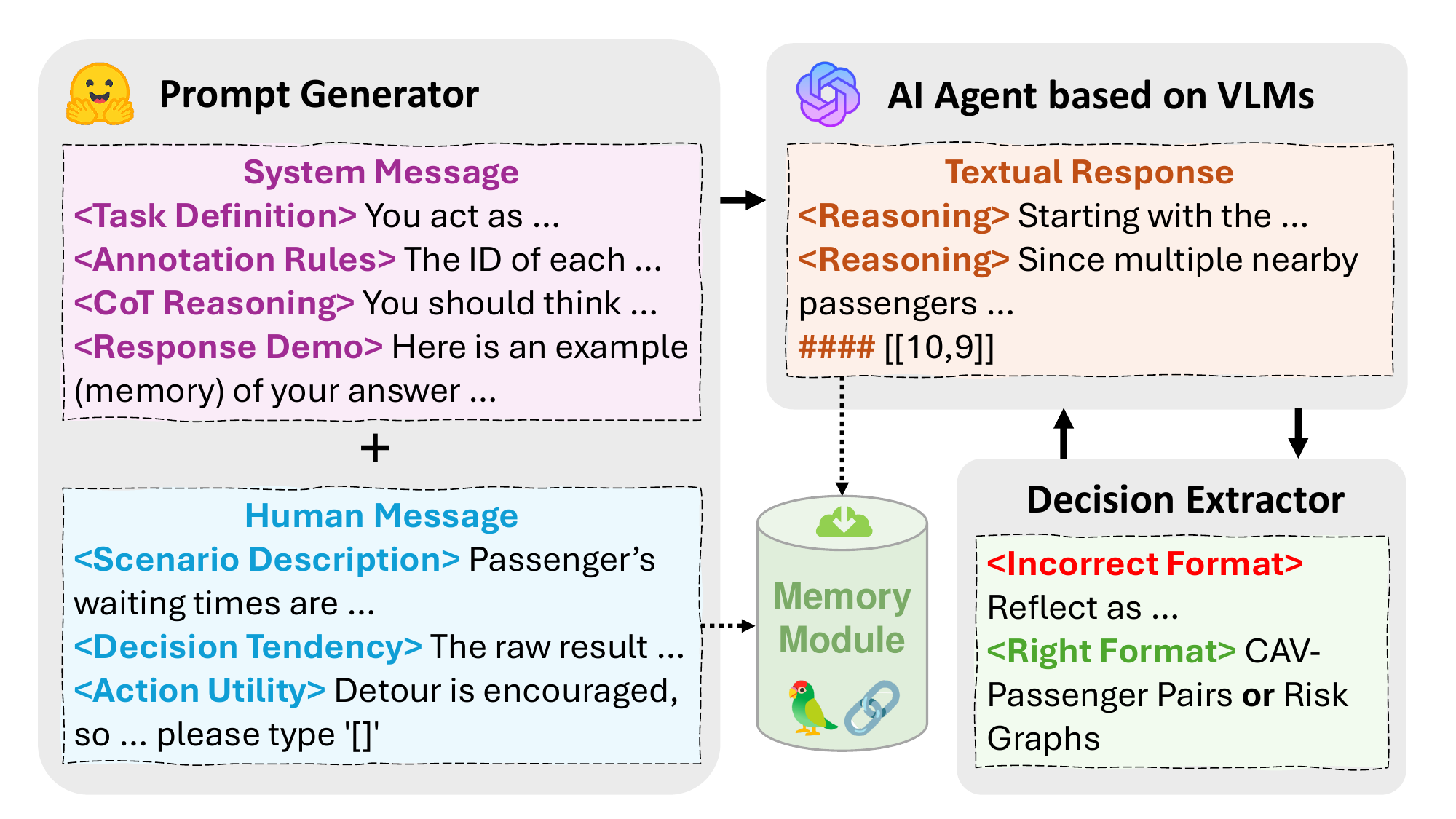}
    \caption{Demonstration of the textual dialogue of the VLM agent. The system message is the abstract of the common knowledge of the environment and the task commands for the AMoD services. The human message enhances the understanding of the corresponding BEV image for the VLM agent by supplementing necessary information. The AI message is the textual response of the VLM agent, which is then stored in the memory module as an important resource for the inference of the VLM agent in a new dialog episode.}
    \label{fig:multimodalMessageDemo}
\end{figure}
\paragraph{Supplementary Text Annotation}
As shown in Fig.~\ref{fig:multimodalMessageDemo}, the language-enhanced map with supplementary text $L(\mathcal{O})$ comprises three distinct components: firstly, a predefined \textit{system message}, which does not pertain to the specific scenario at a given time step; secondly, a \textit{human message} that delineates the status of traffic entities at each time step when invoking the VLM; and thirdly, a \textit{memory message} that aggregates previous similar human messages compared with the current situation, along with the corresponding responses called \textit{AI message} generated by the VLM agent. 
The system message component provides detailed descriptions of the driving environment, including the representation method of different lane marking types, vehicle categorization, and the problem to be solved for the VLM agent. Historical messages stored in the memory module serve as contextual learning references, elaborated in Section~\ref{subsec:memoryModule}. The human message includes additional inputs such as the numerical positions with the IDs of the unresponded requests and the free vehicles in the Cartesian coordinate of the AMoD system.

To optimize the reasoning process and minimize the number of tokens utilized, the distance matrix $\bm{D}$ is provided to the VLM agent. Let $ M $ and $N$ denote the total numbers of free vehicles and existing passenger requests, respectively. $ \bm{D} = [d_{ij}] \in \mathbb{R}^{M \times N} $ captures the distance of vehicles in $ \mathcal{G} $ and passengers in $\mathcal{C}$.  
Mathematically, the distance matrix $\bm{D} = [d_{ij}]$ is defined as:
\begin{equation}\label{eq:distMat}
    \bm{D} = 
    \begin{bmatrix}
    d_{11} & d_{12} & \cdots & d_{1N} \\
    d_{21} & d_{22} & \cdots & d_{2N} \\
    \vdots & \vdots & \ddots & \vdots \\
    d_{M1} & d_{M2} & \cdots & d_{MN} \\
    \end{bmatrix},
\end{equation}
where $d_{ij}$ represents the Euclidean distance between free vehicle $i$ and unresponded passenger request $j$. This matrix is instrumental in determining the optimal assignment of vehicles to passenger requests, as each element $d_{ij}$ is computed using the formula:
\begin{equation}
    d_{ij} = \sqrt{(x_i - x_j)^2 + (y_i - y_j)^2},
\end{equation}
where $(x_i, y_i)$ and $(x_j, y_j)$ are the Cartesian coordinates of vehicle $i$ and passenger request $j$, respectively. This quantitative input aids in formulating efficient routing strategies and optimizing the dispatching process. This matrix quantifies the spatial relationships between vehicles and passenger requests, thereby supplying a quantitative basis for dispatching decisions. By incorporating $\bm{D}$, the VLM agent can more efficiently evaluate distances, which enhances its ability to make informed decisions.
Overall, the multimodal input for the VLM agent, illustrated in Fig.~\ref{fig:bev_image_demo} and Fig.~\ref{fig:multimodalMessageDemo}, integrates essential contextual information from multiple sources, including the road layout, passenger requests, and vehicle states, alongside task commands from the AMoD system. It is also necessary to establish a memory container to facilitate the AI agent's continuous learning and adaptation capabilities. 
\begin{figure}[t]
    \centering
    \includegraphics[width=1\linewidth]{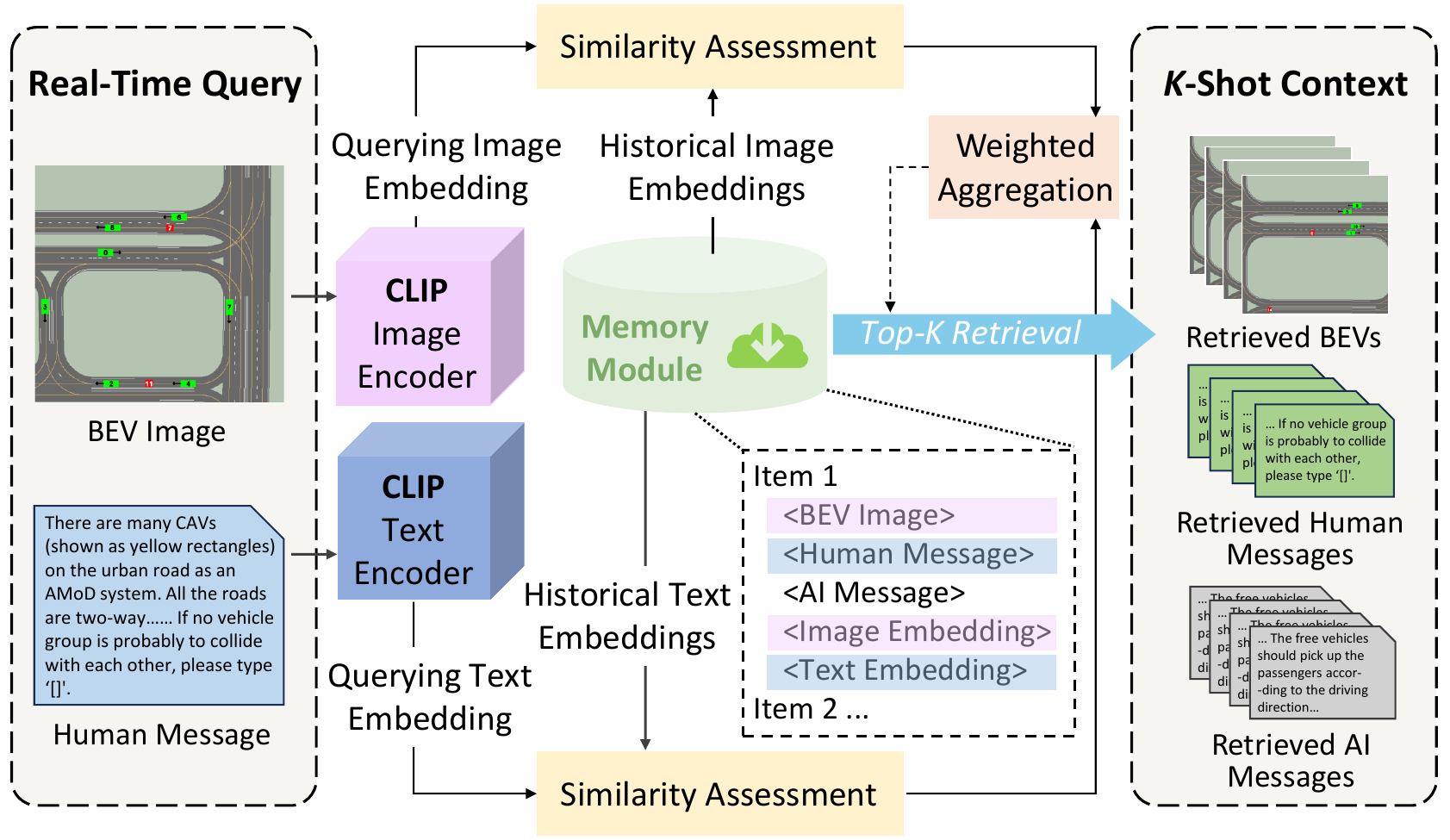}
    \caption{Illustration of the retrieval process for CoDriveVLM. A BEV image and human message for similarity query are embedded with the pre-trained CLIP encoders, before the pair-wise similarity quantification with the memory vectors. The memory items of Top-$K$ similarities are retrieved to support in-context learning.}
    \label{fig:MemoryRetrieve}
\end{figure}
\subsection{Memory Container and Top-K Retrieval}\label{subsec:memoryModule}
\subsubsection{Memory Storage and Organization}
The memory module serves as a repository for multimodal data, storing initialized memory items that include manually annotated responses to guide the AI agent's reasoning process. Each memory item consists of a BEV image, a human message, and the corresponding AI agent response according to the dialogue structure described in Fig.~\ref{fig:multimodalMessageDemo}, forming a comprehensive dataset for in-context learning of the CoDriveVLM.
The reflection module, integrated into the memory system, allows for the iterative refinement of decision-making capabilities. By analyzing previous response errors and their contexts, the system can adapt its strategies and improve its performance in similar future instances.

To enable efficient retrieval of similar dialogues from the memory dataset, we store the embeddings associated with each memory item, as illustrated in Fig.~\ref{fig:MemoryRetrieve}. Specifically, the embeddings of both images and their corresponding human message texts are stored in vector format. This approach facilitates rapid and accurate queries by the retrieval module, as detailed in Section~\ref{subsubsec:retrieveTopK}. The vector representation approach of these embeddings allows for effective similarity measurements, thereby enhancing the retrieval process's performance and efficiency. When it is necessary to retrieve similar memory items, their embeddings are quickly extracted and compared with the embedding of the current instance.

\subsubsection{Retrieval of Top-K Memories}\label{subsubsec:retrieveTopK}
The retrieval module is designed to access the memory module efficiently, enabling the system to draw upon historical data for informed decision-making with in-context learning. When the AI agent produces an incorrect output, the retrieval module identifies and processes these errors, ensuring continuous learning and adaptation.

For effective retrieval, each memory item is assigned an image embedding of the BEV and a text embedding of the human message, generated by pre-trained encoding models, which together represent the multimodal scenario features of the item. We employ the vision encoder and text encoder of CLIP~\cite{radford2021learning} for their validated performance in various semantic comprehension tasks. The overall similarity between the current scenario and each memory item is evaluated using a weighted aggregation of cosine similarities calculated with image embeddings and corresponding text embeddings:
\begin{equation}
    \begin{aligned}
        \text{Sim}(\bm{c}_{\alpha}, \bm{c}_{\beta}) &= \omega \cdot \frac{\bm{o}_\alpha \cdot \bm{o}_\beta}{\|\bm{o}_\alpha\| \|\bm{o}_\beta\|} + (1 - \omega) \cdot \frac{\bm{l}_\alpha \cdot \bm{l}_\beta}{\|\bm{l}_\alpha\| \|\bm{l}_\beta\|},
    \end{aligned}
\end{equation}
where $\bm{o}_\alpha = \mathcal{T}_o(\mathcal{O_\alpha})$ and $\bm{l}_\alpha = \mathcal{T}_l(\mathcal{L_\alpha})$ are the image and text embeddings of the current scenario $\bm{c}_{\alpha}$ respectively, while $\bm{o}_\beta$ and $\bm{l}_\beta$ represent the embeddings of a memory item $\bm{c}_{\beta}$, where $\beta \in \{1,2,...,M\}$. Note that $\mathcal{T}_o(\cdot)$ and $\mathcal{T}_l(\cdot)$ represent the functions generating the embeddings from the corresponding context, and $M$ is the number of the memory items in the container. Besides, $\omega$ is a weighting parameter that can be adjusted to balance the two sources of feature representations, and it is set to 0.5 in this study. Based on the similarity scores, the module retrieves the top-$K$ most similar memories, forming a few-shot-learning context composed of historical BEVs, human messages, and AI messages, which guides the AI agent's reasoning process and enhances the adaptability and effectiveness of the AMoD system's operations.
\subsection{Dispatching Decisions of the VLM Agent}
At each time step when the adaptive frequency system is triggered by~(\ref{eq:scheduleTriggering}), the VLM agent is called for the dispatching assignment of CAVs. To enhance the reliability of the reasoning result, we incorporate the Chain-of-Thought (CoT) reasoning template to the input prompts of the VLM agent as described in the system message of Fig.~\ref{fig:multimodalMessageDemo}. For clear analysis, we formulate the prompting-reasoning process as follows:
\begin{equation}\label{eq:llmInfoFlow}
    \{\mathcal{P}_\tau,\mathcal{C}_\tau\} = \mathcal{T}(K(\mathcal{S},\mathcal{M}_\tau, \mathcal{H}_\tau)),
\end{equation}
where the input elements of the context generator $K(\cdot)$ are constant system message $\mathcal{S}$, memory message $\mathcal{M}_\tau$, and the human message $\mathcal{H}_\tau$ for the current time step $\tau$. Assuming the VLM can simultaneously support continuous dialogue with multimodal context, we denote the function $\mathcal{T}$ as the reasoning process of the VLM agent. Inspired by~\cite{wei2022chain}, the output of the VLM agent is designed as two folds: The final ID pairs $\mathcal{P}_\tau$ of free CAVs and assigned passenger requests, and the text during the CoT reasoning process $\mathcal{C}_\tau$. With the CoT result, we can enhance the transparency of the dispatching decision made by the VLM agent, and it is profoundly beneficial for the reflection module to reflect the decision quality for further enhancement of the memory dataset.

Afterward, $\mathcal{P}_\tau$ is employed as high-level guidance of the cooperative motion planning of the CAVs assigned tasks to pick up and drop off passengers. 
The cooperative motion planning scheme is also enhanced by the same VLM agent structure, which is elaborated in Section~\ref{sec:vlm-admmSystem}.
\section{VLM-ADMM-hybrid System for Vehicle Cooperative Motion Planning}\label{sec:vlm-admmSystem}

The development of AMoD systems necessitates sophisticated strategies for cooperative motion planning and control for the CAVs. This is particularly vital in urban environments where the complexity of traffic dynamics and the presence of multiple autonomous vehicles can lead to route conflicts and inefficiencies.
By leveraging shared information and coordinated implicit decision-making, cooperative motion planning can effectively mitigate the conflicts led by their initial routes, thereby enhancing overall transportation safety. Besides, this collaborative approach also contributes to the smooth flow of traffic, reducing congestion and improving traffic efficiency in densely populated urban areas.
\subsection{Overall Architecture}
To cater to the limited reliable communication range and achieve cooperative motion planning for large-scale CAVs in future AMoD systems, the VLM-ADMM-Hybrid System under the structure of CoDriveVLM is designed through a systematic framework. Central to this system is the on-demand generation of reference paths for each CAV, which allows for real-time adaptability to changing traffic conditions and passenger demands. This capability ensures that vehicles can dynamically adjust their trajectories in response to new information, enhancing their operational efficiency and reliability for passengers.

To implement this system, we formulate an OCP for each subgraph of the CAV network. This formulation is crucial as it defines the objectives and constraints that govern the cooperative driving behavior of the vehicles. By breaking down the larger agent network into manageable subgraphs, we can tailor the motion planning process to the specific conditions and interactions present in each localized segment of the traffic environment. This parallel approach not only simplifies the computational complexity associated with global trajectory optimization but also facilitates the integration of dynamic passenger request assignments, enabling the system to respond swiftly to the AMoD system. Ultimately, the VLM-ADMM-Hybrid system represents a significant advancement in cooperative driving frameworks, providing a robust foundation for the successful deployment of AMoD systems in future urban landscapes.

\subsection{Initial Path Generation for the Assigned CAVs}\label{subsec:init_path_gen}
The foundation of our path planning strategy is to create a connectivity graph $\mathcal{G}^c=(\mathcal{W}^c,\mathcal{E}^c)$ that represents the urban road network. We start by generating waypoints $\mathcal{W}^c$, which are equidistant points positioned along the centerline of each lane as drawn in the right sub-figure of Fig.~\ref{fig:bev_image_demo}. To establish the graph's edges $\mathcal{E}^c$, we link each waypoint to its immediate predecessor and successor within the same lane. We also create connections to corresponding waypoints in the neighboring left and right lanes to account for lane change flexibilities.

First, to generate an initial path for a CAV, we consider its current position and orientation, $\bm{x}^s = [p^{s}_x, p^{s}_y, \varphi^s]^\top$, and its target destination $\bm{x}^t = [p^{t}_x, p^{t}_y, \varphi^t]^\top$ extracted from $\bm d_j\in \bm R_j\in \mathcal{R}$. We employ the A$^*$ algorithm to determine the shortest path on the graph representation $\mathcal{G}^c$. This approach is favored for its computational efficiency and its ability to maintain continuity in the path by utilizing a simplified graph representation of the actual roads.

Second, the resulting path generated by searching in $\mathcal{G}^c$, denoted as $\hat{\bm{P}} = [\bm{x}^s, \bm{x}^1, \bm{x}^2, \ldots, \bm{x}^t] \in \mathbb{R}^{3 \times p}$, where $p$ is the number of waypoints in the path, serves as an initial reference for CAV navigation. To refine this path for enhanced spatial-temporal motion planning, we first assess the Cartesian distance $d^{a,a+1}$ between each pair of adjacent waypoints $\bm{x}^a$ and $\bm{x}^{a+1}$. If this distance exceeds the product $v_{\text{ref}} \cdot \Delta T \cdot w$, additional waypoints are interpolated along the segment to ensure smoother transitions. Note that $v_{\text{ref}}$ is the reference velocity, $\Delta T$ is the time step, and $w_\text{dis}=1.2$ is a slack factor.

Third, the path's continuity and smoothness are improved using a Savitzky-Golay filter~\cite{chen2004simple}, resulting in the refined path $\bm{P} = f_{\text{sg}}(\hat{\bm{P}})$. This filtering process ensures an even distribution of waypoints, particularly in critical areas such as intersections, corners, and zones requiring lane changes. Following the establishment of global reference waypoints $\bm{x}^a \in \bm{P}$, each CAV updates its local reference trajectory by employing a nearest neighbor search enabled by KD-Tree~\cite{ram2019revisiting}. At each time step, the CAV identifies the nearest waypoint from the waypoints in its initial path $\bm{P}^i$, where $i$ denotes the CAV's unique identifier.
Through this comprehensive path generation and refinement process, the CoDriveVLM enhances the effectiveness of CAV navigation and supports real-time adaptability in dynamic urban settings. 
\subsection{VLM-based Collision Risk Evaluation for Subgraphs}\label{subsec:}

Before cooperatively optimizing the initial paths of the CAVs, collision risk evaluation is critical for ensuring safe and efficient navigation of the CAVs. Its necessity arises from two folds: First, the objective limitations of V2X communication, which is stable only within a limited range. Second, incorporating excessive collision avoidance constraints can significantly degrade computational efficiency. Therefore, it is imperative to focus on essential constraints to optimize the sets of trajectories individually and parallelly.
Collusion risk evaluation is intricately linked to the generation of subgraphs that formulate the OCPs. Vehicles identified as having potential collision risks are grouped into a single subgraph, where their collision avoidance constraints are collectively managed during the motion planning procedure.

The pipeline for subgraph generation involves several key stages, starting with multimodal input generation. The input includes a BEV image featuring all active CAVs, their driving directions, and IDs, highlighted in sky-blue rectangles similar to the right sub-figure of Fig.~\ref{fig:bev_image_demo}. Supplementary text annotations provide additional information, including data annotated by the VLM dispatching agent in Section~\ref{subsubsec: annotationVLMDispatchingAgent}, as well as current and reference velocities for each CAV. Furthermore, the design of textual information is a critical aspect. This comprehensive input facilitates a more informative and accurate evaluation of collision risks. 
Following input generation, a reasoning template is applied to process the information using a CoT approach as shown in~Fig.~\ref{fig:multimodalMessageDemo}. In each sub-image of the BEV, all CAVs are analyzed in the context of their surrounding vehicles, considering factors such as driving directions, distances, and relative speeds. The analysis results in the identification of subgroups containing vehicles with potential collision risks, which are then summarized and outputted as textual information using the format of (\ref{eq:llmInfoFlow}).

The decoding of the VLM agent's response is the final stage in the pipeline. The response, structured with delimiters, includes a list of lists representing the CAV IDs. 
Each inner list indicates the CAVs suggested for inclusion in the same subgraph for cooperative trajectory optimization. 
Edges are created between each pair of CAVs within these lists, forming an updated graph $\mathcal{G}_\tau$ at time step $\tau$. Linked vehicles are grouped into the same subgraph $\mathcal{H}_{k,\tau}$, corresponding to the $k$th OCP for the planning horizon starting at $\tau$ for AMoD.

Overall, this VLM-based collision risk evaluation framework divides the CAVs into smaller-scale subgraphs and effectively enhances the cooperative motion planning process by focusing on essential collision avoidance constraints, thereby improving computational efficiency in the AMoD system. 
\subsection{Parallel Optimization for CAVs in Each Subgraph}\label{subsec:ParallelOpt}
For the scheduled AMoD vehicle fleet, we consider a cooperative motion planning task within subgraph $\mathcal{H}$ generated by the VLM-guided graph evolution module. Each vehicle $ n^i \in \mathcal{H} $ has a state vector $ \bm{z}^i_\tau $ and a control input vector $ \bm{u}^i_\tau $ at each time step $\tau$. We formulate this task, involving $N_s$ CAVs with specified destinations and paths $\bm P^i, i\in \{0,1,...,N_s-1\}$ generated in Section~\ref{subsec:init_path_gen}, as an OCP:
\begin{equation}\label{eq:NMPCOptProb}
\begin{array}{ll}
\underset{{\bm{z}_\tau^i}, {\bm{u}_\tau^i}}{\min} & \sum_{i=1}^N Q_i(\bm Z^i, \bm U^i)\\
\text { s.t.} & \boldsymbol z^i_{\tau+1}=f(\boldsymbol z^i_\tau,\boldsymbol u^i_\tau),  \\
& \bm z^i_{\tau+1}\in \mathcal{S}^i_\tau,\\
& -\boldsymbol {\underline u}^i \preceq \boldsymbol {u}_\tau^i \preceq \boldsymbol {\overline u}^i,\\
% & -\boldsymbol {\underline z}^i \preceq \boldsymbol z_\tau^i \preceq \boldsymbol {\overline z}^i,\\
% & -\boldsymbol {\underline z}^i \preceq \boldsymbol z_T^i \preceq \boldsymbol {\overline z}^i,\\
&\forall \tau \in \mathcal{T}_p,\ \forall i \in \mathcal{N}_s,
\end{array}
\end{equation}
where $\mathcal{N}_s$ represents the set of indices of CAVs within a subgraph $\mathcal{H}_s$, and $\mathcal{T}_p = \{0, 1, \ldots, T_p-1\}$ denotes the planning horizon for a cooperative motion planning episode. The kinematic model of the CAVs is denoted as $f(\boldsymbol z^i_\tau,\boldsymbol u^i_\tau)$, and $\mathcal{S}_\tau^i$ defines the collision-free and bounded operational region for the $i$th CAV at time step $\tau$. ${\bm{\underline{u}}}^i$ and ${\bm{\overline{u}}}^i$ represent the lower and upper bounds of the control input for the $i$th CAV, respectively. The objective of this OCP is to enable the CAVs to follow their reference paths assigned by the upstream scheduling, which can be expressed as:
\begin{equation}\label{eq:individualObj_t}
    Q_i(\bm z^i_\tau, \bm u^i_\tau) = \sum _{\tau=0}^T \| \bm z^i_\tau-\bm z^i_{\text{ref},\tau}\|^2_{\bm Q}+ \sum_{\tau=0}^
{T-1}\| \bm u^i_\tau\|^2_{\bm R},
\end{equation}
where $\bm Q$ and $\bm R$ are the diagonal weighting matrices balancing the path tracking and energy saving. Assume the state and control input of the vehicles in the kinematic model are $\bm s_\tau=[p_{x,\tau},p_{y,\tau},\varphi_\tau,v_\tau]^\top$ and $\bm u_\tau = [a_\tau, \delta_\tau]^\top$, respectively, the concurrent reference $\bm z_{\text{ref},\tau}$ is queried as described in the third step of Section~\ref{subsec:init_path_gen} with a default constant reference velocity $v_\text{ref}=10$\, m/s. With the notation of $$\bm Z^i=[\bm z_0^i, \bm z_1^i, ... , \bm z_{T_p}^i],\bm U^i=[\bm u_0^i, \bm u_1^i, ... , \bm u_{T_p-1}^i],$$we concatenate the temporal series objectives in~(\ref{eq:individualObj_t}) as $Q_i(\bm Z^i, \bm U^i) = \sum_{\tau=0}^{T_p-1}Q_i(\bm z^i_\tau, \bm u^i_\tau)$. At this stage, the OCP for the subgraph $\mathcal{H}$ of CAVs is fully defined. 
The next subsection then addresses parallel and cooperative planning, building on an individual OCP to achieve integrated solutions.
\begin{figure}[t]
    \centering
    \includegraphics[width=1\linewidth]{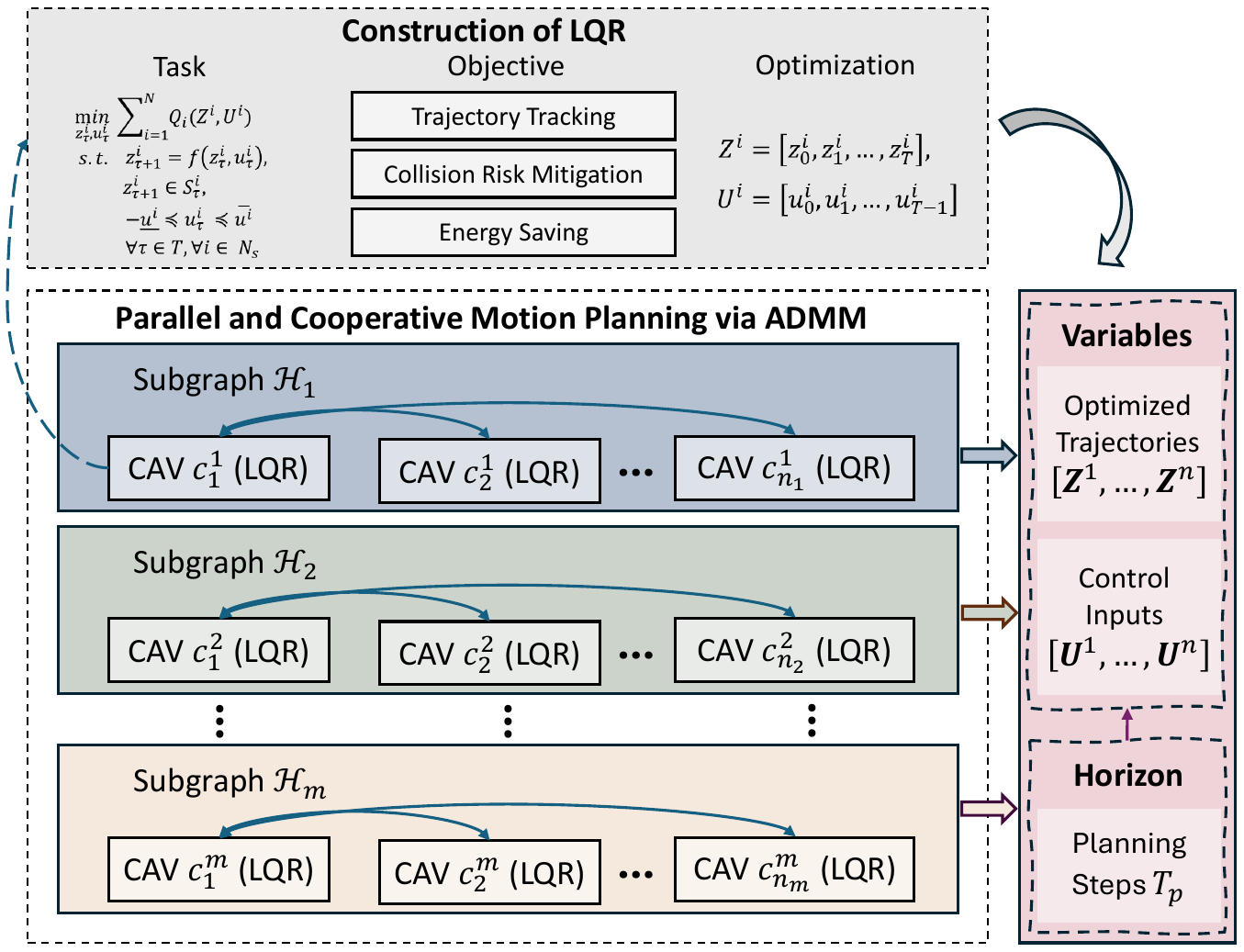}
    \caption{The structure of the parallel and cooperative motion planning algorithm. There is no information exchange between any two sets of subgraphs $\mathcal{H}_1$ and $\mathcal{H}_2$, while the CAVs in each subgraph need to communicate with others iteratively.}
    \label{fig:parallelCooperativeStructure}
\end{figure}

\begin{algorithm}[htbp]
\caption{Cooperative Dispatching and Motion Planning for CAVs in the AMoD System}\label{alg:alg3}\label{alg:alg2}
\begin{algorithmic}[1]
\State \textbf{While} {simulation time $<T_\text{sim}$ \textbf{or} $\exists$ passenger is not arrived at its destination}
\State \textbf{schedule} target positions of all CAVs in $\mathcal{G}$ using VLM dispatching agent
\State \textbf{determine} optimal route using A* algorithm for each CAV from current state to the destination
\State \textbf{smooth} the waypoints in each route by the Savitzky-Golay filter
\State \textbf{While} {$\exists$ requests in the AMoD}
\State Execute VLM-guided graph evolution to distribute the CAVs to $\mathcal{H}_i$
\State Initialize \texttt{ProcessPoolExecutor} with a pool of workers
\State \textbf{For} {subgraph $\mathcal{H}_k$ in subgraphs list $\{\mathcal{H}_1, \mathcal{H}_2, \cdots \mathcal{H}_n\}$}
\State Submit task to \texttt{processPool}:
\State \hspace{0.3cm} \textbf{search} nearest waypoints for each CAV at each time step $\tau \in \{0,1,...,T_s\}$ using KD-tree of the initial path $\bm P$ in the optimal routes
\State \hspace{0.3cm} \textbf{For} the CAVs within $\mathcal{H}_k$:
\State \hspace{0.3cm} \textbf{initialize} $\{x^i_\tau,u^i_\tau\}^{T_s}_{\tau=0},$
\State \hspace{1.72cm} $\{p^{i,0},y^{i,0},z^{i,0},s^{i,0}\}, \forall i\in \mathcal{N}_k$
\State \hspace{0.3cm} \textbf{choose} $\sigma,\rho >0$
\State \hspace{0.3cm} \textbf{repeat}:
\State \hspace{0.6cm} Send $\{z^i_\tau\}_{\tau=1}^T$, receive $\{z^j_\tau\}_{\tau=1}^T$ from $j\in\mathcal{N}^i_k$
\State \hspace{0.6cm} Compute $k^i$, $J^i$, $\{A^i_\tau\}^{T-1}_{\tau=0}$, $\{B^i_\tau\}^{T-1}_{\tau=0}$
\State \hspace{0.6cm} \textbf{reset} $p^{i,0}=s^{i,0}=0, y^{i,0}=y^\text{last},x^{i,0}=x^\text{last}$
\State \hspace{0.6cm} \textbf{reset} $y^{i,0}_{[2]} = x^{i,0}_{[2]}=0$
\State \hspace{0.6cm} \textbf{repeat}: for all $i \in \mathcal{N}_k$
\State \hspace{0.9cm} Send $y^{i,k}$, receive $y^{j,k}$ from $j\in\mathcal{N}^i_k$
\State \hspace{0.9cm} Broadcast $y^{i,k}$ to the vehicles in $\mathcal{N}^i_k$
\State \hspace{0.9cm} $p^{i,k+1}=p^{i,k}+\rho\sum_{j\in \mathcal{N}^i_k}(y^{i,k}-y^{j,k})$
\State \hspace{0.9cm} $s^{i,k+1}=s^{i,k}+\sigma(y^{i,k}-x^{i,k})$
\State \hspace{0.9cm} $r^{i, k+1} = \sigma x^{i, k}+\rho \sum_{j\in \mathcal{N}^i_k}\left(y^{i, k}+y^{j, k}\right)$
\Statex \hspace{1.9cm} $-(k^i+p^{i, k+1}+s^{i, k+1})$
\State \hspace{0.9cm} Perform (\ref{subeq:imp1})-(\ref{subeq:imp3}) for $ \alpha^{i,k}_{[2,i]}, \alpha\in\{ p, s,  r\}$
\State \hspace{0.9cm} Perform (\ref{eq:sub1})-(\ref{eq:sub3}) for $ \alpha^{i,k}_{[2,j]},\alpha\in\{ p, s,  r\}$,
\Statex \hspace{5.4cm} $j\in\mathcal{N}^i_k$
\State \hspace{0.9cm} Compute $X^{i, k+1}$ by solving LQR problem (\ref{eq:quadOptProb})
\State \hspace{0.9cm} $y^{i, k+1}_{[1]} = 2\gamma\left(\hat J^i X^{i, k+1}_{[1]}+r^{i, k+1}_{[1]}\right)$
\State \hspace{0.9cm} ${x}_{[1]}^{i, k+1} = \Pi_{\mathcal{R}^{\circ}_+}\left(\frac{1}{\sigma}s^{i, k+1}_{[1]}+ y^{i, k+1}_{[1]}\right)$
\State \hspace{0.9cm} Perform (\ref{subeq:imp4}-\ref{subeq:imp5}) for $ \alpha^{i,k}_{[2,i]}, \alpha\in\{ y, x\}$
\State \hspace{0.9cm} Perform (\ref{eq:sub4}-\ref{eq:sub5}) for $\alpha^{i,k}_{[2,j]}, \alpha\in\{ y,x\},j\in\mathcal{N}^i_k$
\State \hspace{0.9cm} $k=k+1$
\State \hspace{0.6cm} \textbf{until} number of iteration steps exceeds $k_\text{max}$
\State \hspace{0.6cm} Update $\{z^i_\tau,u^i_\tau\}^T_{\tau=0}$
\State \hspace{0.3cm} \textbf{EndFor}
\State \hspace{0.3cm} \textbf{until} termination criterion is satisfied
\State \textbf{EndFor}
\State \texttt{processPoolExecutor} waits for all tasks to complete
\State \textbf{perform} first $T_p$ steps of the CAVs in $\mathcal{G}$
\State \textbf{feedback} at time step $T_p$
\State \textbf{relay} the states of all the CAVs in $\mathcal{G}$
\State \textbf{EndWhile}
\State \textbf{EndWhile}
\end{algorithmic}
\end{algorithm}
\subsection{Parallel and Cooperative Motion Planning via ADMM}

As illustrated in Fig.~\ref{fig:parallelCooperativeStructure}, the CAVs in different $\mathcal{H}_k$ do not exchange any information during the whole planning horizon $T_p$, which manner is fully paralleled. On the other hand, the CAVs within the same $\mathcal{H}$ have their own LQR problem with consensus variables with other CAVs. The consensus variables are updated between iterations with necessary communication. This part introduces the construction of the LQR problems and the iterative consensus updating for the whole system using dual consensus ADMM~\cite{liu2024improved}.
In order to reduce the size of the OCP and provide an agile response, MPC is used with a planning horizon $T_p$, and an executing horizon $T_e$. Before giving the dual formulation of the OCP of (\ref{eq:NMPCOptProb}), it is convexified and reformulated by the following procedures. First, the collision avoidance constraints in 
$\bm z^i_{\tau+1}\in \mathcal{S}^i_\tau$ is locally linearized with Gauss-Newton approximation at each time step, and the quadratic objective function is reformulated by second-order Taylor expansion to cater to the perturbed form of the independent variables $\Delta \bm Z^i$ and $\Delta \bm U^i$. With its Lagrangian, the dual formulation of the OCP is as follows:
\begin{equation}
    \begin{aligned}
        \min_{\bm {\Delta X}^1,...,\bm{\Delta X}^{N_m}} &\sum_{i=1}^{N_m} F^i(\bm {\Delta X}^i)+\mathcal{I}_{\mathcal{K}}\left(\bm h\right)\\
        \text{s.t. } & \sum_{i=1}^{N_m} \left(\bm J^i\bm {\Delta X}^i-\bm k^i\right)=\bm h,
    \end{aligned}
\label{ConstrainedNewProb}
\end{equation}
where $\bm{\Delta X}^i=\left[\Delta \bm X^i_0;\Delta \bm X^i_1;\cdots;\Delta\bm X^i_{T_p-1}\right]$ with $\Delta\bm X^i_{\tau} = \left[\Delta \bm Z^i(\tau);\Delta \bm U^i(\tau)\right]$. Let $F^i$ denote the hosting cost for CAV $i$, and $\mathcal{I}_\mathcal{K}$ represent the indicator function for mutual avoidance and bounded constraints.
Based on the dual consensus ADMM algorithm proposed in~\cite{grontas2022distributed}, leveraging the Lagrangian of (\ref{ConstrainedNewProb}) with the dual variables ${\bm p, \bm s, \bm r,\bm y,\bm x}$, the cooperative motion planning of the vehicles in the AMoD system is derived as shown in Algorithm~\ref{alg:alg2}. It contains two main processes, \textit{dual update} for the dual variables using steps 22-27 and steps 29-32, and \textit{primal update} to explore the optimal states and control inputs assuming the variables of other vehicles remain constant using step 28 in Algorithm~\ref{alg:alg2}.

For the dual update process, the variables $\bm x^{i,k}_{[2,i]}$ belonging to the target vehicle $i$ at the $k$th iteration in the individual optimization problem are updated by
\begin{subequations}\label{eq:dualUpdate_tv}
    \begin{align}
\label{subeq:imp1}
\bm{p}_{[2, i]}^{i, k+1} & =\bm{p}_{[2, i]}^{i, k}+\rho(N-1)\left(\bm{y}_{[2, i]}^{i, k}-\bm{y}_{[2, i]}^k\right), \\ 
\label{subeq:imp2}
\bm{s}_{[2, i]}^{i, k+1} & =\bm{s}_{[2, i]}^{i, k}+\sigma\left(\bm{y}_{[2, i]}^{i, k}-\bm{x}_{[2, i]}^{i, k}\right), \\ 
\bm{r}_{[2, i]}^{i, k+1} & =\sigma \bm{x}_{[2, i]}^{i, k}+\rho(N-1)\left(\bm{y}_{[2, i]}^{i, k}+\bm{y}_{[2, i]}^k\right)\notag\\
\label{subeq:imp3}
&-(\bm{k}_{[2, i]}^{i, k+1}+\bm{p}_{[2, i]}^{i, k+1}+\bm{s}_{[2, i]}^{i, k+1}), \\
\label{subeq:imp4}
\bm{y}_{[2, i]}^{i, k+1} & =2\gamma\left(\bm O^i \bm{\Delta X}^{i, k+1}+\bm{r}_{[2, i]}^{i, k+1}\right), \\
\label{subeq:imp5}
\bm{x}_{[2, i]}^{i, k+1} & =\Pi_{\mathcal{S}^{b\circ}}\left(\frac{1}{\sigma}\left(\bm{s}_{[2, i]}^{i, k+1}+\bm{y}_{[2, i]}^{i, k+1}\right)\right) .
\end{align}
\end{subequations}
On the contrary, in order to reduce the number of constraints and the amount of communication data, the updating of the variables $\bm x^{v,k}_{[2,i]}$ for the surrounding linked vehicles $v\in \mathcal{N}^i$ in the same LQR problem is executed as follows:
\begin{subequations}\label{eq:dualUpdate_sv}
    \begin{align}
\label{eq:sub1}
\bm{p}_{[2, i]}^{v, k+1} & =\bm{p}_{[2, i]}^k+\rho\left(\bm{y}_{[2, i]}^k-\bm{y}_{[2, i]}^{i, k}\right) = \bm{p}_{[2, i]}^{k+1}, \\
\label{eq:sub2}
\bm{s}_{[2, i]}^{v, k+1} & =\bm{s}_{[2,1]}^k+\sigma\left(\bm{y}_{[2, i]}^k-\bm{x}_{[2, i]}^k\right)=\bm{s}_{[2, i]}^{k+1}, \\
\bm{r}_{[2, i]}^{v, k+1} & =\sigma \bm{x}_{[2, i]}^k+\rho(2 |n^v|-1) \bm{y}_{[2, i]}^k+\rho \bm{y}_{[2, i]}^{i, k}\notag  \\
&\label{eq:sub3}
-(\bm{k}_{[2, i]}^{k+1}+\bm{p}_{[2, i]}^{k+1}+\bm{s}_{[2, i]}^{k+1})=\bm{r}_{[2, i]}^{k+1},\\
\label{eq:sub4}
\bm{y}_{[2, i]}^{v, k+1} & =2\gamma \bm{r}_{[2, i]}^{k+1}=\bm{y}_{[2, i]}^{k+1}, \\
\label{eq:sub5}
\bm{x}_{[2, i]}^{v, k+1} & =\Pi_{\mathcal{S}^{b\circ}}\left(\frac{1}{\sigma}\left(\bm{s}_{[2, i]}^{k+1}+\bm{y}_{[2, i]}^{k+1}\right)\right)=\bm{x}_{[2, i]}^{k+1}.
\end{align}
\end{subequations}
Note that for the dual update of $\bm{y}_{[2, i]}^{v, k+1}$ for surrounding vehicles connected to the target vehicle, only consensus with $\bm{r}_{[2, i]}^{k+1}$ in (\ref{eq:sub4}) is necessary. In contrast, the target vehicle must also account for box constraints via high-dimensional matrix multiplication in (\ref{subeq:imp4}), a process that is computationally intensive and may not add significant value.

For the primal update, owing to the fully decentralized nature of the algorithm, the standard LQR problem considers only the independent variables in $\bm{\Delta X}^i$ as follows:
\begin{equation}
\begin{aligned}
    \min_{\bm {\Delta X^i}}\, &\frac{1}{2}\bm{\Delta X}^{i\top}\bm L^i_2\bm{\Delta X}^i + 2\bm r^{i,k+1\top}\bm J^i\bm{\Delta X}^{i}\\
    &+\bm{\Delta X}^{i\top}\bm L^i_1+\frac{1}{\sigma+2\rho |n^i|}\bm{\Delta X}^{i\top}\bm J^{i\top}\bm J^i\bm{\Delta X}^{i}  \\
    \text{s.t. }\, &\left(\bm L^i_3-\bm L^i_4\right) \bm{\Delta X}^i=0,
\end{aligned}
\label{eq:quadOptProb}
\end{equation}
where $\bm J^i$ is the Jacobian matrix for collision risk mitigation with surrounding vehicles. Besides, $\bm L_1^i$ and $\bm L_2^i$ are the first-order and second-order partial derivatives of the objective function , while $\bm L_3^i-\bm L_4^i$ constitute the local gradient of the vehicle kinematic model in (\ref{eq:NMPCOptProb}). 
In conclusion, as shown in Algorithm~\ref{alg:alg3}, during task execution, vehicles evolve the connection graph using the VLM graphing agent and solve cooperative motion planning problems in parallel and iteratively through a receding horizon approach until all the passengers have arrived at their destinations.

\section{Experimental Results and Analysis}
\subsection{Environment Setup}
\begin{figure}[htbp]
    \centering
    \includegraphics[width=1\linewidth]{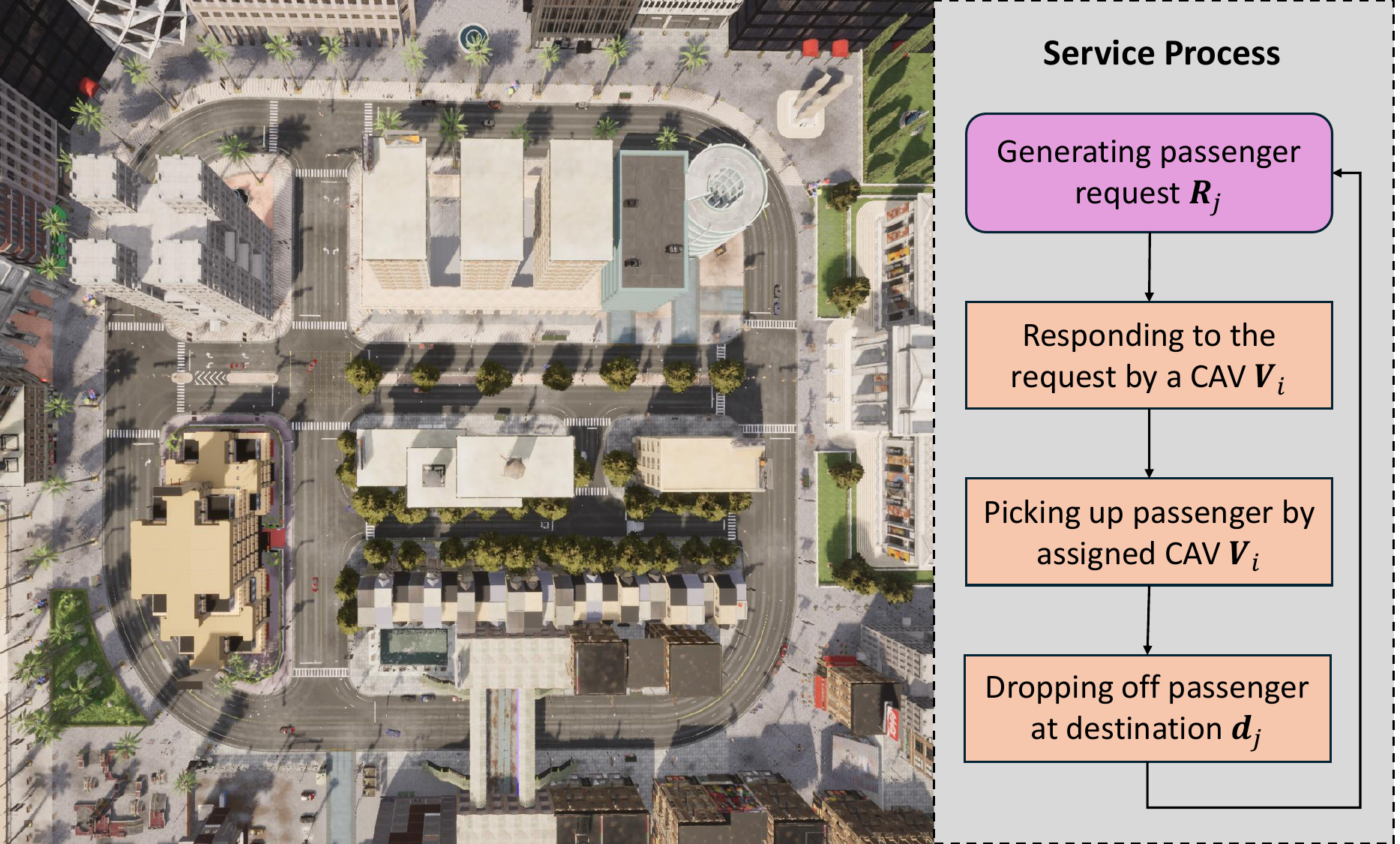}
    \caption{Demonstration of the environmental settings of the urban scenario and the corresponding service process flow of the AMoD System.}
    \label{fig:ExperimentalSettings}
\end{figure}
To evaluate the efficacy of the proposed cooperative dispatching and motion planning framework, we conduct experiments in the following computational environment. The implementation is carried out on a system running Ubuntu 20.04 LTS, equipped with an Intel(R) Xeon(R) 8358P CPU operating at 2.60\,GHz. The computational capabilities for the neural networks are enhanced by NVIDIA RTX 4090 GPU with $24\times 4$\,GB of graphic memory, spread across four units. The system's memory infrastructure includes 16 channels, each with 64\,GB of RAM, providing ample resources for intensive data processing of the large-scale and complex cooperative driving tasks in the presented AMoD system.
For the software environment, we utilize Python 3.8 with PyTorch. To optimize the performance of numerical computations, particularly those involving heavy data manipulation and iteration, we employ Numba, a Just-In-Time compiler for Python that translates a subset of NumPy code into fast machine code.

The simulation experiments are conducted on the CARLA simulator, version 0.9.14, a widely used open-source platform designed for autonomous driving research~\cite{dosovitskiy2017carla}. CARLA provides a realistic urban driving environment, crucial for testing and validating autonomous vehicle algorithms. Specifically, our experiments are set in \texttt{Town10}, one of CARLA's pre-defined urban layouts, which offers a complex road network ideal for evaluating the proposed framework's capabilities in dynamic and challenging scheduling and planning scenarios, as illustrated in Fig.~\ref{fig:ExperimentalSettings}. 
It is essential to delineate that a complete service cycle in the context of this study is defined by a sequence of stages: the generation of a passenger request, the acknowledgment and response to this request by a CAV, the subsequent pickup of the passenger by the CAV, and the final drop-off at the predetermined destination. The experimental evaluation meticulously examines these stages to assess the overall performance of the proposed CoDriveVLM framework and varieties of baselines. The foundation model used in the experiments is GPT-4o accessed by Azure.

The kinematic behavior of CAVs is modeled using the non-linear bicycle model as described in \cite{tassa2014control}. The vehicle's state vector $\bm z = [x, y, \theta, v]^\top$ and control input vector $\bm u$ are denoted in Section~\ref{subsec:ParallelOpt}. Note that the heading angle relative to the positive $x$-axis is given by $\theta$, and $\delta$ is the steering angle of the front wheel.
The vehicle's kinematic model is mathematically described by:
\begin{equation}
\label{eq:bicycle_dynamics}
\begin{split}
\bm z_{\tau+1} = f(\bm z_\tau, \bm u_\tau) = \left[\begin{array}{c}
x_\tau + f_i(v_\tau, \delta_\tau) \cos \theta_\tau \\
y_\tau + f_i(v_\tau, \delta_\tau) \sin \theta_\tau \\
\theta_\tau + \arcsin \left(\frac{g(v_\tau, \delta_\tau)}{b_i}\right) \\
v_\tau + \tau_s a_\tau
\end{array}\right]
\end{split},
\end{equation}
where the function $g(v_\tau, \delta_\tau) = v_\tau \Delta T \sin(\delta_\tau)$, and $f_i(v, \delta)$ is defined as:
\begin{equation}
f_i(v, \delta) = b_i + v \Delta T \cos \delta - \sqrt{b_i^2 - g(v, \delta)^2},
\end{equation}
with $b_i$ representing the wheelbase of the $i$th CAV and $\Delta T$ is the discrete time step. Additional parameters related to the experiments are presented in Table~\ref{tab:expParams}.
% Table generated by Excel2LaTeX from sheet 'Parameters'
\begin{table}[t]
  \centering
  \caption{Parameter Settings in Simulations}
    \begin{tabular}{lccc}
    \toprule
    \multicolumn{2}{c}{Parameter} & Description & Value \\
    \midrule
    \multirow{3}[2]{*}{Kinematics} &  $b$    & Wheelbase of the vehicles & 2.8\,m \\
          &  $l$    & Length of the vehicles & 4.7\,m \\
          & $w$     & Width of the vehicles & 1.8\,m \\
    \midrule
    \multirow{5}[2]{*}{Time-related} & $T_\text{sim}$ & Simulation duration & 200 s \\
          & $T_s$  & Request updating period & 10\,s \\
          & $T_p$ &  Trajectory planning horizon & 2\,s \\
          & $\Delta T$    & Discrete time step & 0.1\,s \\
          & $T_\text{max}$  & Maximum waiting time & 40\,s \\
    \bottomrule
    \end{tabular}%
  \label{tab:expParams}%
\end{table}%

\subsection{Comparative Study}
% Table generated by Excel2LaTeX from sheet 'comparison of TR and TC'
\begin{table*}[htbp]
  \centering
  \caption{Comparison of task response and task completion for CoDriveVLM and the baselines with different CAV scale and passenger request numbers and phases}
    \begin{tabular}{cccccccc}
    \toprule
    \multicolumn{2}{c}{Env. Settings} & \multirow{2}[2]{*}{Metric} & \multicolumn{5}{c}{Method} \\
    Veh. count & Req. count &       & CoDriveVLM & Dist. First & Idle First & FCFS  & Mix. First \\
    \midrule
    \multirow{4}[2]{*}{10} & \multirow{4}[2]{*}{30} & $T_{atr}$ (s) &  $\bm{61.44 \pm 28.12}$     & $74.33 \pm 40.08$ &   $ 79.16 \pm 47.55$    &   $66.00 \pm 23.05$    & $84.00 \pm 38.43$ \\
          &       & $T_{atc}$ (s) &    $\bm{91.22 \pm 30.96}$   & $98.80 \pm 32.35$ &   $ 94.00 \pm 44.65$    &    $102.14 \pm 29.96$    &  $108.94 \pm 27.15$\\
          &       & Response Rate &  $60.0\%$     & $60.0\%$    &    $63.3\%$   &    $53.3\%$   & $\bm{70.0\%}$ \\
          &       & Completion Rate &   $\bm{60.0\%}$    & $50.0\%$    &    $50.0\%$   &    $46.7\%$   & $56.7\%$ \\
    \midrule
    \multirow{4}[2]{*}{15} & \multirow{4}[2]{*}{30} & $T_{atr}$ (s) &   $62.41 \pm 37.45$    &    $65.29 \pm 37.46$   &    $86.92 \pm 54.28$   &  $\bm{58.55 \pm 35.12}$     & $68.07 \pm 33.62$ \\
          &       & $T_{atc}$ (s) &   $94.71 \pm 44.53$    &   $90.07 \pm 52.25$    &    $ 111.52 \pm 45.20$   &  $\bm{83.24 \pm 40.45}$     & $93.76 \pm 43.47$ \\
          &       & Response Rate &   $\bm{96.7\%}$    & $93.3\%$  &   $86.7\%$    &  $73.3\%$     & $93.3\%$ \\
          &       & Completion Rate &   $\bm{93.3\%}$    &  $86.7\%$  &   $70\%$    &   $70.0\%$    & $83.3\%$ \\
    \midrule
    \multirow{4}[2]{*}{15} & \multirow{4}[2]{*}{40} & $T_{atr}$ (s) &   $58.29 \pm 39.22$    &   $59.87 \pm 41.67$    &    $75.45 \pm 50.87 $   &    $ \bm{56.09 \pm 36.38}$   & $81.24 \pm 48.45$ \\
          &       & $T_{atc}$ (s) &    $82.30 \pm 42.23$   &   $ \bm{74.96 \pm 45.84}$    &   $ 89.22 \pm 45.08$    &    $78.63 \pm 37.54 $   & $ 96.78 \pm 48.71$ \\
          &       & Response Rate &    $\bm{85.0\%}$   &    $75.0\%$  &   $72.5\%$    &   $55.0\%$    & $72.5\%$ \\
          &       & Completion Rate &   $\bm{67.5\%}$    &    $57.5\%$   &   $57.5\%$    &   $47.5\%$    &  $57.5\%$\\
    \midrule
    \multirow{4}[2]{*}{20} & \multirow{4}[2]{*}{40} & $T_{atr}$ (s) &    $53.14 \pm 43.41$   & $\bm{52.34 \pm 39.14}$     & $60.00 \pm 40.84$    & $58.90 \pm 40.47$      & $69.66 \pm 35.64$  \\
          &       & $T_{atc}$ (s) &    $94.19 \pm 42.92$   & $\bm{89.53 \pm 40.49}$     &  $110.76 \pm 43.84$    & $95.83 \pm 40.75$    & $108.64 \pm 36.74$ \\
          &       & Response Rate &   $\bm{92.5\%}$    & $87.5\%$    & $80.0\%$      & $72.5\%$      & $87.5\%$  \\
          &       & Completion Rate &   $\bm{80.0\%}$    &  $75.0\%$   & $72.5\%$    & $60.0\%$     & $70.0\%$ \\
    \midrule
    \multirow{4}[2]{*}{25} & \multirow{4}[2]{*}{40} & $T_{atr}$ (s) &   $35.00 \pm 34.74$    &  $\bm{33.03 \pm 28.45}$     &  $69.66 \pm 35.64$     & $53.50 \pm 32.98$    & $57.94 \pm 32.82$ \\
          &      & $T_{atc}$ (s) &  $\bm{63.06 \pm 44.15}$     & $63.33 \pm 35.33$      & $108.64 \pm 36.74$     & $94.21 \pm 40.56$      & $98.18 \pm 39.10$ \\
          &       & Response Rate &  $\bm{90.0\%}$     &  $87.5\%$     & $87.5\%$   & $80\%$     & $\bm{90.0\%}$ \\
          &       & Completion Rate &   $\bm{85.0\%}$    &  $82.5\%$     & $70.0\%$    & $72.5\%$      & $\bm{85.0\%}$ \\
    \bottomrule
    \end{tabular}%
  \label{tab:comparisonAllBaselines}%
\end{table*}%
We assess the effectiveness of the proposed CoDriveVLM scheme alongside several representative baselines, including Distance First (Dist. First), Idle First, First Come First Serve (FCFS), and Mixed First (Mix. First). This evaluation is conducted within the constructed AMoD system by measuring a range of performance indicators and qualitatively illustrating representative instances.
The Dist.~First method employs a distance matrix as described in (\ref{eq:distMat}) to schedule CAVs by identifying the closest CAV-request pairs in ascending order, thereby facilitating efficient pairings. In contrast, the Idle~First method prioritizes available vehicles that have been idle for the longest duration, assigning them to the nearest passengers in sequence. Conversely, the FCFS method adopts a passenger-centric approach, prioritizing those passengers who have waited the longest. Finally, the Mix.~First method primarily utilizes the distance matrix $\bm D(t)$ for scheduling; however, if there are passengers whose waiting time exceeds a predefined threshold $t - T_{sp,j} > T_\text{max}$ shown in Table~\ref{tab:expParams}, priority is shifted to the passenger $j$ and the closest CAV is assigned to serve this request for better user experience.

\begin{figure}[h]
    \centering
    \includegraphics[width=1\linewidth]{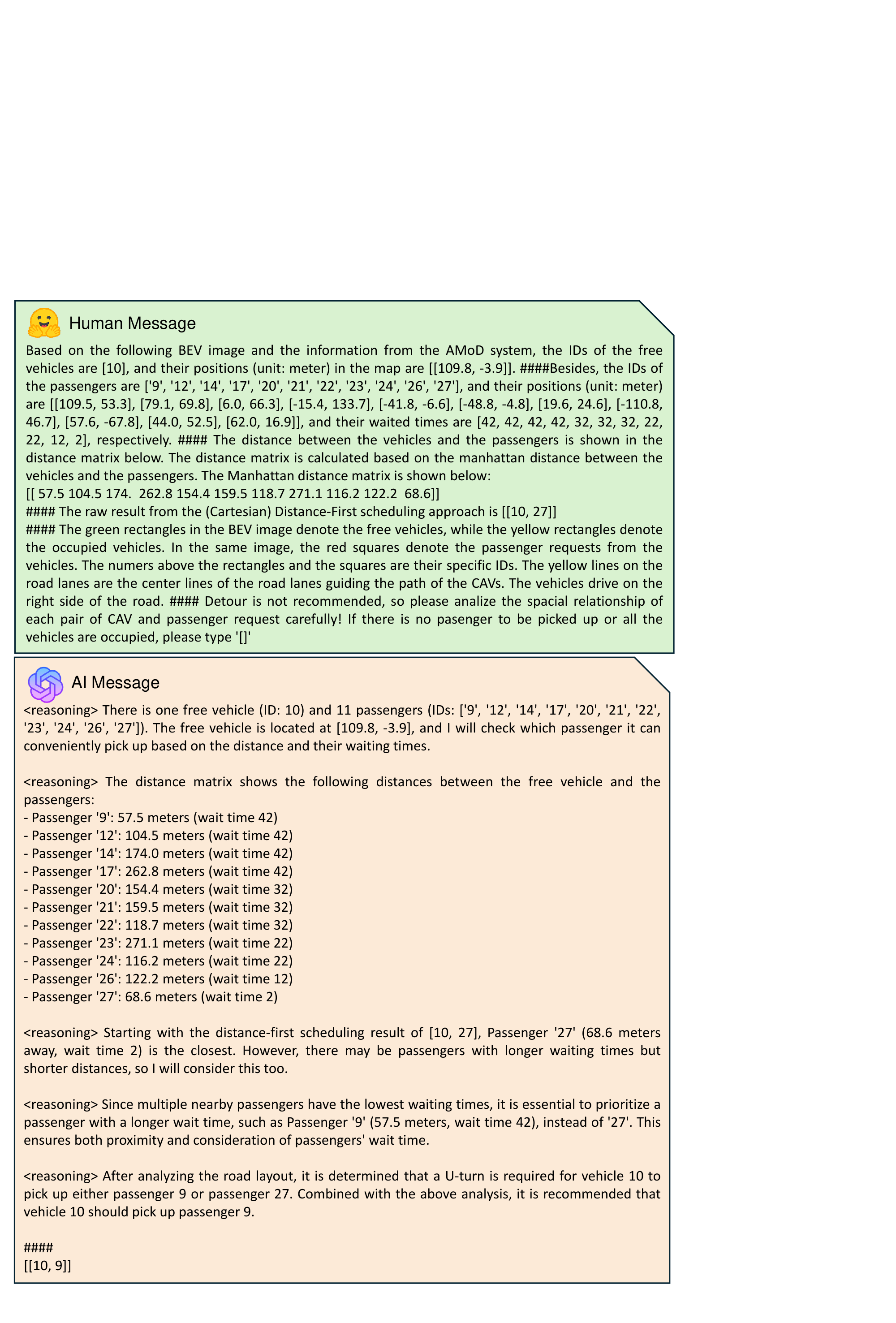}
    \caption{The corresponding dialogue for the dispatching decision at the traffic condition at 42.0\,s in the simulation trial with 15 CAVs and 30 predefined passenger requests. The AI agent of the CoDriveVLM inferences and reasons with CoT to get the decision of assigning CAV 10 to pick up passenger 9.}
    \label{fig:qualitative_dialogue2}
\end{figure}

To evaluate the overall dispatching performance, the indices we utilize are two service rates and their corresponding time spent in each task, namely response rate $RR(t)$, completion rate $CR(t)$, average task response time (or waiting time) $T_{atr}$, and average task completion time $T_{atc}$. The response rate $ RR(t) = \left( \frac{N_R(t)}{N_T} \right) \times 100\%$ at a specific time step $ t $ quantifies the proportion of travel requests that have been successfully serviced by the fleet up to that time step. It is calculated as the ratio of the number of requested passengers that have been picked up to a vehicle $N_R(t)$ and completed by time $t$, to the total number of requests $N_t$ received by the system. Similar definition is applicable to $ CR(t) = \left( \frac{N_C(t)}{N_T} \right) \times 100\%$. In the simulation results presented in Table~\ref{tab:comparisonAllBaselines}, the simulation duration is set to \( t = T_\text{sim} = 200 \, \text{s} \), which contains time steps $\tau\in\{0,1,...,1999\}$ with $\Delta T=0.1$\,s. On the other hand, in terms of the individual task service quality evaluation, average task response time $T_{atr}$ is utilized to evaluate the reasonableness of the dispatching decisions, while the average task completion time $T_{atc}$ is vital to assess the performance of cooperative motion planning during the service. The definitions are as follows:
\begin{equation}
    T_{atr} = \frac{1}{M}\sum_{j=1}^M (T_{\text{pk},j} - T_{\text{sp},j}),
\label{eq:Tatr}
\end{equation}
\begin{equation}
    T_{atc} = \frac{1}{M}\sum_{j=1}^M (T_{\text{ar},j} - T_{\text{sp},j}),
\label{eq:Tatc}
\end{equation}
where $M$ is the number of responded passenger requests, and $T_{\text{pk},j}$, $T_{\text{pk},j}$, and $T_{\text{pk},j}$ share the same notation in Section~\ref{subsec:AmodStructure}.
As illustrated in Table~\ref{tab:comparisonAllBaselines}, we conduct closed-loop simulations of cooperative dispatching and motion planning across five distinct environmental settings for a comprehensive assessment. Each setting features a total of 10 to 25 vehicles and 30 to 40 passenger requests to simulate the traffic conditions at different time of a day.
Due to the traffic condition perception and road layout understanding via the BEV image and the supplementary textual information, our proposed CoDriveVLM achieves the best response rate and completion rate in most of the environment settings, which means the overall dispatching efficiency for the whole traffic system is remarkable. Due to the heuristic rules of the FCFS method, its average task response time is the best in the environment settings of $15\times 30$ and $15 \times 40$, while the Distance First method achieves the best average task completion performance in the environment settings of $15\times 40$ and $20 \times 40$. Note that the Mixed First method achieves the most efficient response rate in the $10\times 30$ environment setting.
\begin{figure*}[t]
\centering
\subfigure[10 CAVs with 30 passengers at 146\,s]{
\includegraphics[ width=0.31\linewidth]{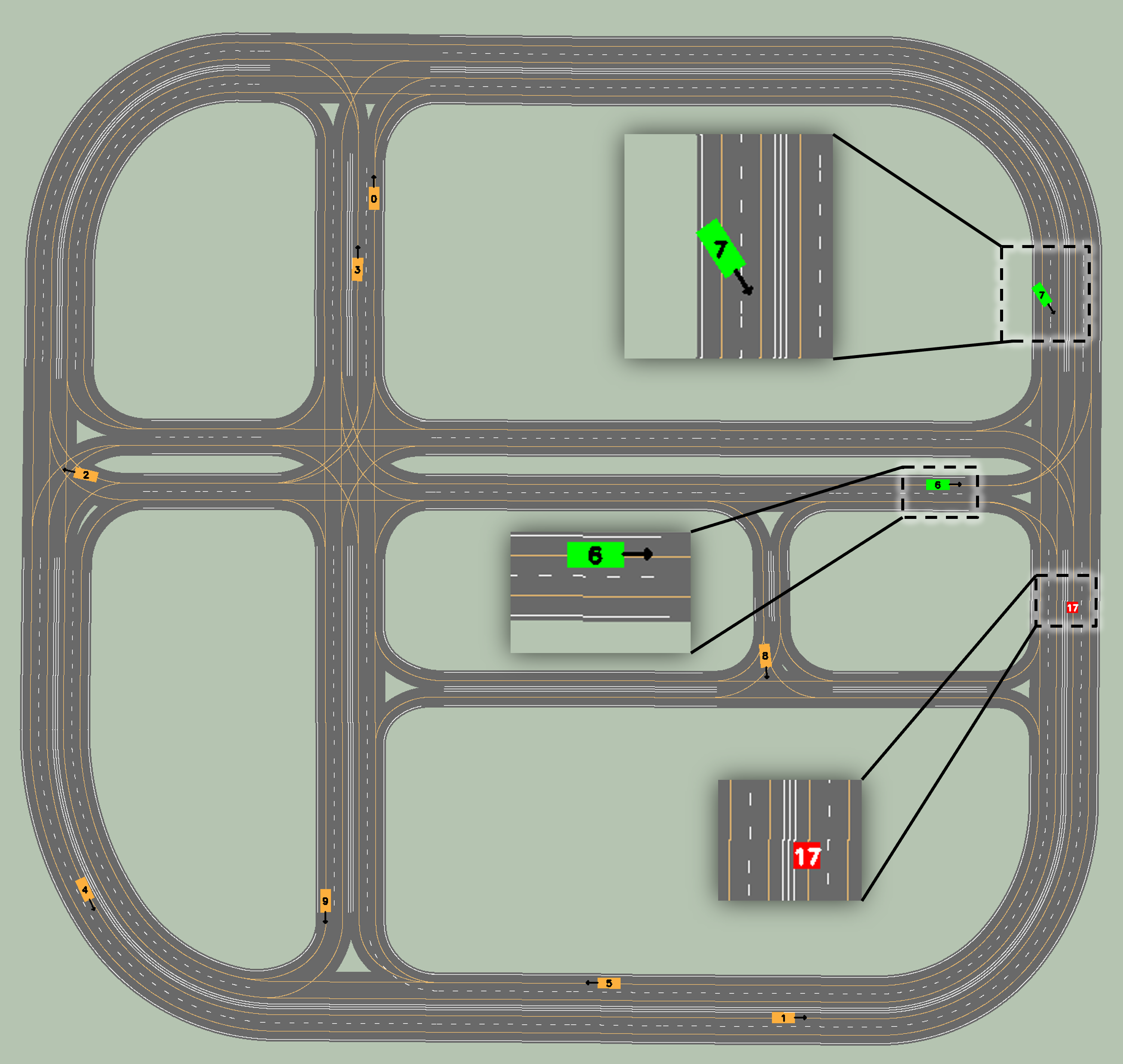}
\label{subfig:qualitativeDemo1}
}
\subfigure[15 CAVs with 30 passengers at 42\,s]{
\includegraphics[width=0.31\linewidth]{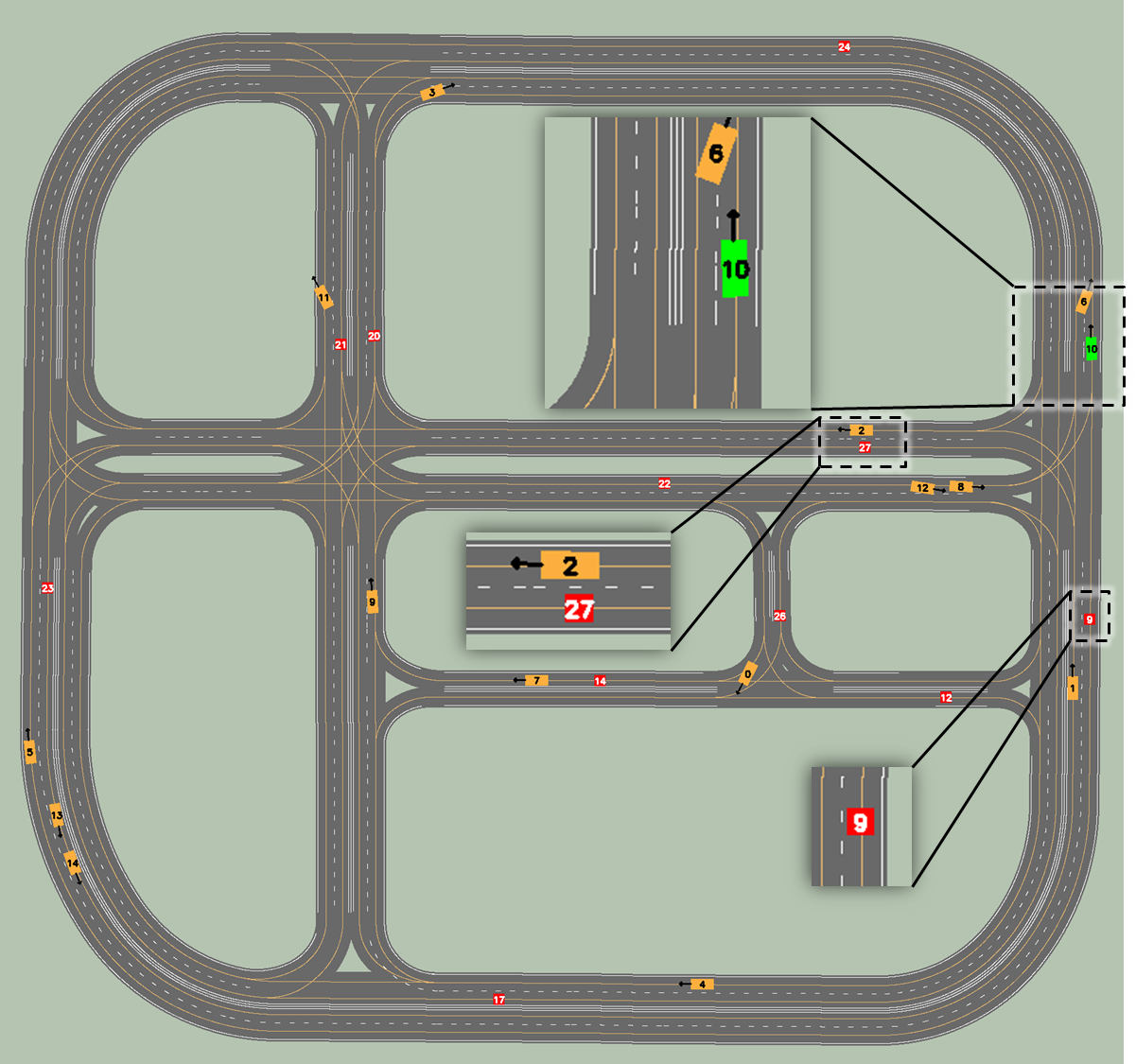}
\label{subfig:qualitativeDemo2}
}
\subfigure[20 CAVs with 40 passengers at 60\,s]{
\includegraphics[width=0.31\linewidth]{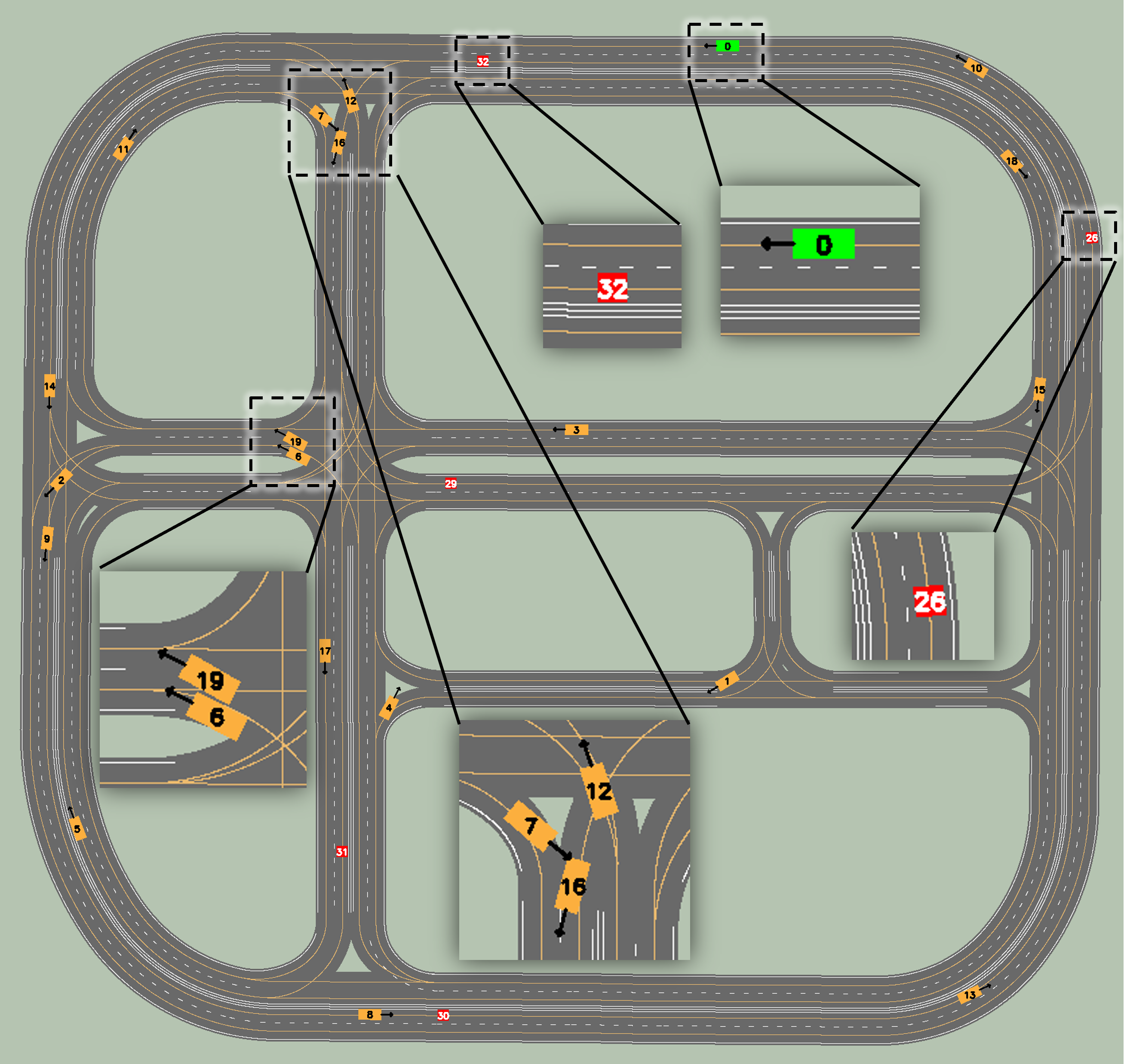}
\label{subfig:qualitativeDemo3}
}
% \DeclareGraphicsExtensions.
\caption{The traffic conditions described by the integrated BEV images in different simulation trials using our proposed CoDriveVLM method. Key elements in the BEV images have been enlarged, respectively.}
\label{fig:qualitativeDemo}
\end{figure*}
\begin{figure}[h]
    \centering
    \includegraphics[width=1\linewidth]{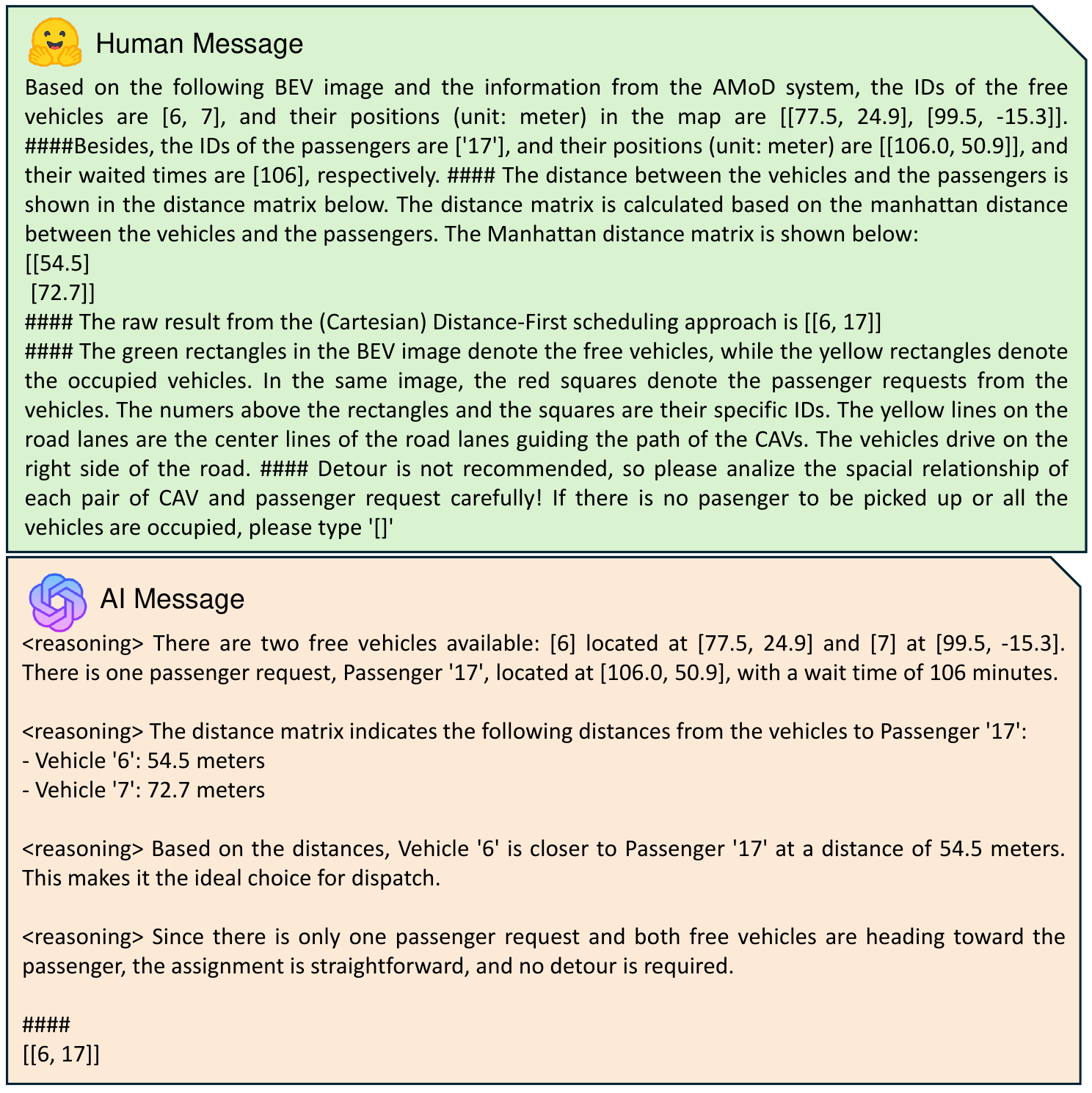}
    \caption{The corresponding dialogue for the dispatching decision at the traffic condition at 146.0\,s in the simulation trial with 10 CAVs and 30 predefined passenger requests. The AI agent of the CoDriveVLM inferences and reasons with CoT to get the decision of assigning CAV 6 to pick up passenger 17.}
    \label{fig:qualitative_dialogue1}
    \vspace{-1.3em}
\end{figure}
% \vspace{-1em}
Representative instances demonstrating the effectiveness of the CoDriveVLM are provided in Fig.~\ref{fig:qualitativeDemo}. Specifically, Fig.~\ref{subfig:qualitativeDemo1} illustrates the dispatching condition at 146.0\,s with the environment setting of $10\times 30$, and the key elements are enlarged for better understanding of the reasoning and inferencing process of the AI agent in our proposed CoDriveVLM method. In this case, there are two CAVs and only one request on the map, indicating that most of the existing passenger requests are assigned in the previous operating process. Fig.~\ref{fig:qualitative_dialogue1} illustrates how the corresponding human message provides essential contextual information. This information includes the positions of each traffic participant, the distances between each CAV and passenger 17, and the passenger's waiting time. Since there is only one passenger request at this time step, the AI agent disregards the waiting time and primarily focuses on distance and spatial relationships. On the contrary, Fig.~\ref{subfig:qualitativeDemo2} shows the BEV image incorporating the traffic condition at 42.0\,s in the $15\times 30$ environment setting. In this condition, there is only one free vehicle, while there are 11 passenger requests awaiting response. For a focused and concise reasoning process of the AI agent, the Manhattan distances from vehicle 10 to all the unassigned passenger requests are provided using textual information as demonstrated in the human message of Fig.~\ref{fig:qualitative_dialogue2}. In addition, for a better user experience of the passengers, the list of waiting time of the passengers is provided as an important information resource when considering multiple choices to assign. Lastly, the annotation rules of the BEV image as VLM inputs are also described with language information for a better understanding of the spatial-related traffic conditions. Note that the clue of ``detour is discouraged'' is also provided in the last sentence of the human message to be sent to the AI agent to explore the potential of the foundation model's spatial understanding ability. Moreover, based on the above prompt, the AI agent gives a response in a CoT as follows: the instances of free vehicles and passenger requests, the integration of the distances and waiting time for the corresponding vehicle-passenger pairs, the reflection of the initial dispatching result, the reasoning of the dispatching alternatives considering comprehensive indicators, and lastly the final decision of the dispatching pair(s) at the current time step. Specifically, in this situation, the AI agent considers three main factors, distance, waiting time, and necessary driving behavior to pick up each candidate passenger. Although the distance from vehicle 10 to passenger 27 is similar to that to passenger 9, and both require a U-turn for pickup, the AI agent selects passenger 9 due to the longer waiting time of 42.0\,s. This instance reflects the flexibility and reliability of the AI agent to understand the traffic conditions and make reasonable decisions after the CoT process for decision-making for the CAVs in the AMoD systems.

\begin{figure}[t]
    \centering
    \includegraphics[width=0.9\linewidth]{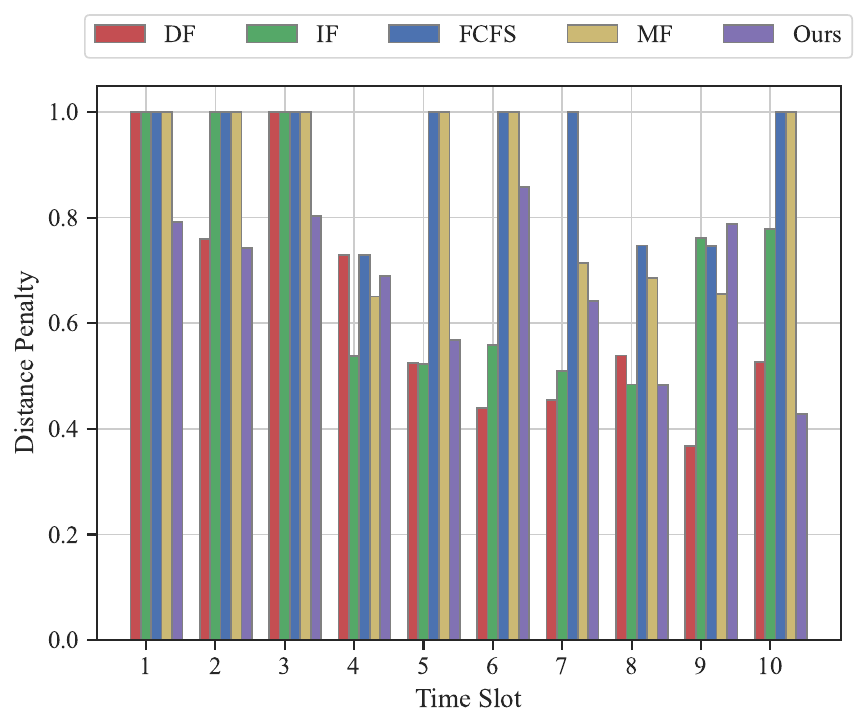}
    \caption{The distance penalty during the cooperative dispatching and motion planning with receding horizon. A smaller distance penalty indicates better safety performance and a smaller collision risk between any pair of CAVs within the time slot.}
    \label{fig:distancePenalty}
\end{figure}

In addition to the traffic efficiency evaluation of the proposed method, the safety performance is also evaluated using distance penalty defined as follows:
\begin{equation}
    DP_k = e^{-\tilde{d_k}}, \tilde{d_k}\in\left\{x\in\mathbb{R}\mid 0<x\leq1\right\}, k\in \mathbb{Z_+},
\end{equation}
where $\tilde{d_k}$ is the normalized minimal distance within the $k$th time slot from $\left(20\cdot (k-1)\right)$\,s to $\left(20\cdot k\right)$\,s until the last time step of the simulation. The mathematical expression of $\tilde{d_k}$ is
\begin{equation}
    \tilde{d_k} = \min \left(\frac{d^{i,j}_\tau}{d_\text{max}}\right), i,j\in \mathcal{N}, \tau\in \mathcal{T}_k
\end{equation}
where $\mathcal{T}_k= \{ x \in \mathbb{R} \mid 200 \cdot (k-1) \leq x < 200 \cdot k \}$ is the time slot in the $k$th interval of the horizontal axis in Fig.~\ref{fig:distancePenalty}. In short, the presented distance penalty $DP_k$ evaluates the minimal distance between pairs of CAVs in a subtle way. According to the characteristic of the exponential function, a smaller $d^{i,j}_\tau$ corresponds to a larger distance penalty. Note that if the minimal distance $\tilde{d}_k$ at the $k$th time slot is close to 0, the distance penalty will be close to $1.0$ indicating a large collision possibility. With the above discussion, from the distance penalty illustrated in Fig.~\ref{fig:distancePenalty}, our proposed CoDriveVLM scheme shows the best safety performance for the small distance penalty without any value close to 1.0 along all the time slots. This result is due to the AI agent used in our graph evolution framework for the parallel OCPs, which can evaluate the spatial relationship between any pair of CAVs in the provided BEV image and then divide the CAVs with collision risks into the same subgraphs. For the baselines using conditional Manhattan distance presented in~\cite{liu2024improved}, they generate trajectories with small distances due to overlooking some of the collision-risk pairs. Therefore, necessary collision avoidance constraints are not incorporated in the corresponding OCPs, resulting in possible collisions.

\subsection{Ablation Study}
The ablation study is conducted by ablating the features of the BEV images, and the few-shot number of memory messages in the prompt of the CoDriveVLM framework. Besides, representative foundation models including ChatGPT 3.5, GPT-4o-mini, Gemini 1.5 Pro,and GPT-4o, are evaluated in the proposed framework to assess the effectiveness of the structure of the CoDriveVLM framework. 

\begin{figure}[t]
    \centering
    \includegraphics[width=0.9\linewidth]{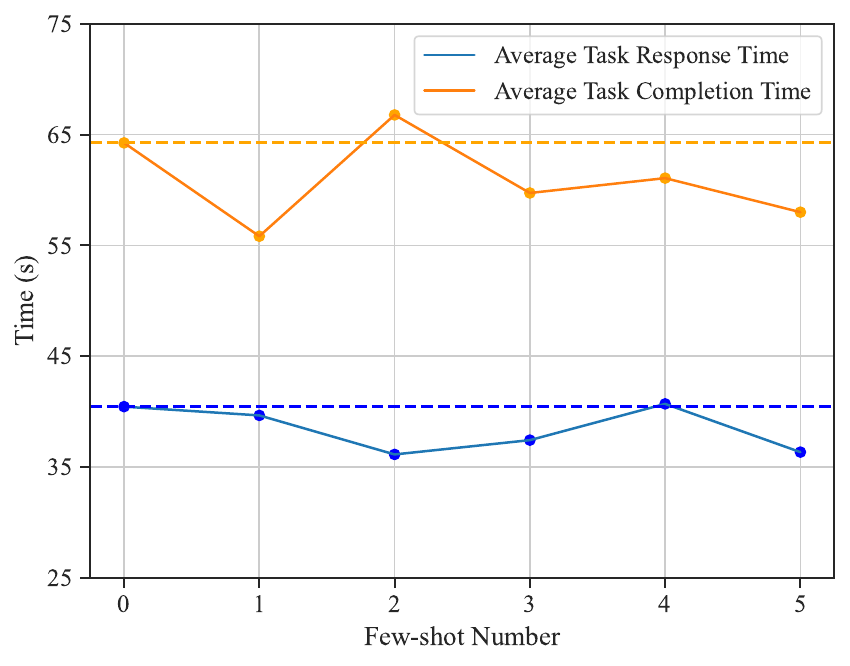}
    \caption{The impact of the memory module and the few-shot number to the performance of the proposed CoDriveVLM framework. The horizontal line is the benchmark value under the zero-shot condition.}
    \label{fig:ablation}
\end{figure}

As shown in Fig.~\ref{fig:ablation}, the task response time and completion time under the condition from zero-shot to five few-shot memories are evaluated. In most cases, the performance of CoDriveVLM with few-shot messages performs better than the zero-shot condition. 
Notably, when the few-shot number is three, the proposed method demonstrates exceptional overall performance. Additionally, we observe that increasing the few-shot number does not necessarily improve the AI agent's performance.
This may be attributed to the increased number of memory fragments, which can lead to confusion within the VLM or induce erroneous reconstructions based on past data. This suggests that an overload of memory fragments might not only impair the model's accuracy but also result in misleading interpretations derived from historical information.

% Table generated by Excel2LaTeX from sheet 'comparison of TR and TC'
\begin{table}[t]
  \centering
  \caption{The cooperative dispatching performance with different foundation models with 20 vehicles and 40 passenger requests}
  \resizebox{1.0\linewidth}{!}{
    \begin{tabular}{ccccc}
    \toprule
    Foundation Model & $T_{atr}$ & $T_{atc}$ & Res. Rate & Comp. Rate \\
    \midrule
    GPT-4o-mini   &   $37.44 \pm 25.97$    &  $59.73 \pm 24.10$     &  $62.5\%$     & $37.5\%$ \\
    Gemini 1.5 Pro   &    $42.43 \pm 24.42$   &  $64.12 \pm 22.60$     &    $70.0\%$   & $40.0\%$ \\
    GPT-4o w/o BEV &   $37.48 \pm 25.24$    &  $65.27 \pm 23.61$     &    $57.5\%$   & $27.5\%$ \\
    GPT-4o  &    $ 35.04 \pm 24.51$   &    $64.71 \pm 23.22$   &    $67.5\%$   &  $35.0\%$ \\
    \bottomrule
    \end{tabular}%
    }
  \label{tab:ablation}%
\end{table}%
Besides, considering that foundation models are the original drivers of AI agents, and their understanding and reasoning abilities significantly affect the performance of the CoDriveVLM framework, we deploy mainstream large multimodal models in the system for a comprehensive evaluation. As illustrated in Table~\ref{tab:ablation}, Gemini 1.5 Pro achieves leading performance in average task completion time, response rate (Res. Rate), and completion rate (Comp. Rate) indicators, while GPT-4o achieves the best average task response performance. Notice that GPT-4o-mini has a comparative performance with the full version of GPT-4o in this mission. We also ablated the BEV image sent to the AI agent of the VLMCoDrive framework demonstrating in the last two rows of Table~\ref{tab:ablation}, which proves the necessity of the multimodal input for the AI agent of the CoDriveVLM framework. Compared to the version without the prompt of BEV image, our approach shows overwhelming advantages in all the indicators.

\section{Conclusion}
Our proposed CoDriveVLM framework represents a significant advancement in the evolution of AMoD systems by tackling the challenges of dynamic and personalized dispatching, optimal routing, and collision risk mitigation. This comprehensive framework for cooperative dispatching and motion planning in CAV systems takes advantage of graph-based representations and VLM-enabled AI agents. Specifically, we innovatively integrate well-annotated BEV images and supplementary textual prompts as inputs to the VLMs. Techniques such as in-context learning and CoT are utilized to explore the AI agent's potential in traffic dispatching and collision risk evaluation for the CAVs. Additionally, leveraging the sparse features of OCPs formulated with initial paths and subgraphs provided by the AMoD platform, we apply consensus ADMM to achieve parallel optimization of CAVs through iterative LQR, with limited information exchange between neighbors. Simulation results confirm that the CoDriveVLM framework effectively enhances traffic efficiency while ensuring safety. The ablation study underscores the necessity of using multimodal prompts for the AI agent. Thus, our cooperative dispatching and motion planning scheme, along with the simulation platform, is beneficial for deployment in future intelligent transportation systems. Part of the future work is to explore the utilization of the proposed cooperative framework in shared AMoD systems with high-capacity vehicles supporting carpools.
\bibliographystyle{ieeetr}
\bibliography{refs}

\end{document}